%% file: Preprint.tex
\documentclass[11pt]{article}

\usepackage[preprint]{acl}

\usepackage{times}
\usepackage{latexsym}

\usepackage[T1]{fontenc}
\usepackage[utf8]{inputenc}

\usepackage{microtype}

\usepackage{inconsolata}
\usepackage{amsmath}
\usepackage{booktabs} 
\usepackage{multirow}
\usepackage{array}
\usepackage[table]{xcolor}
\usepackage{amssymb}
\usepackage{amsthm}
\usepackage[most]{tcolorbox}

\usepackage{graphicx}
\usepackage{xspace}

\newcommand{\eg}{\emph{e.g.,}\xspace}
\newcommand{\ie}{\emph{i.e.,}\xspace}

\newcommand{\wrt}{\emph{w.r.t}\xspace}

\newcommand{\babya}{\textsc{Aslec-drop}\xspace}
\newcommand{\babyb}{\textsc{Aslec-casl}\xspace}
\definecolor{lightgrayv}{HTML}{F4F3F8} % EEEDF3
\definecolor{myblue}{RGB}{82,143,173}
\definecolor{myred}{RGB}{231,98,84}
\title{On the Step Length Confounding in LLM Reasoning Data Selection}

\author{\textbf{Bing Wang}$^{1,2,3}$, \textbf{Rui Miao}$^{3,4}$, \textbf{Chen Shen}$^{3}$\thanks{Corresponding authors}, \textbf{Shaotian Yan}$^{3}$, \textbf{Kaiyuan Liu}$^{3,5}$, \textbf{Ximing Li}$^{1,2,7*}$, \\ 
\textbf{Xiaosong Yuan}$^{1,2,3}$\textbf{, Sinan Fan}$^{3,5}$\textbf{, Jun Zhang}$^{6}$\textbf{, Jieping Ye}$^{3}$ \\
\normalsize \textsuperscript{1} College of Computer Science and Technology, Jilin University \\
\normalsize \textsuperscript{2} Key Laboratory of Symbolic Computation and Knowledge Engineering, MoE, Jilin University \\
\normalsize \textsuperscript{3} Alibaba Cloud Computing \ 
\textsuperscript{4} School of Artificial Intelligence, Jilin University \\
\normalsize \textsuperscript{5} College of Computer Science and Technology, Zhejiang University \\
\normalsize \textsuperscript{6} Department of Mathematics, University of Michigan \
\textsuperscript{7} RIKEN Center for Advanced Intelligence Project \\
\small \texttt{\{wangbing1416,zjushenchen,liximing86\}@gmail.com, miaorui24@mails.jlu.edu.cn}
}

\begin{document}
\maketitle
\begin{abstract}

\input{S_Abstract}

\end{abstract}

\input{S_Introduction}

\input{S_Preliminary}

\input{S_Method}

\input{S_Experiment}

\input{S_RelatedWorks}

\input{S_Conclusion}

\bibliography{reference}

\input{S_Appendix}

\end{document}

%% file: S_Abstract.tex
Large reasoning models have recently demonstrated strong performance on complex tasks that require long chain‑of‑thought reasoning, through supervised fine‑tuning on large‑scale and high‑quality datasets. To construct such datasets, existing pipelines generate long reasoning data from more capable Large Language Models (LLMs) and apply manually heuristic or naturalness‑based selection methods to filter high‑quality samples. 
Despite the proven effectiveness of \textit{naturalness‑based data selection}, which ranks data by the average log probability assigned by LLMs, our analysis shows that, when applied to LLM reasoning datasets, it systematically prefers samples with longer reasoning steps (\ie more tokens per step) rather than higher‑quality ones, a phenomenon we term \textbf{step length confounding}. Through quantitative analysis, we attribute this phenomenon to low-probability first tokens in reasoning steps; longer steps dilute their influence, thereby inflating the average log probabilities.
To address this issue, we propose two variant methods: \babya, which drops first‑token probabilities when computing average log probability, and \babyb, which applies a causal debiasing regression to remove the first tokens’ confounding effect. Experiments across four LLMs and five evaluation benchmarks demonstrate the effectiveness of our approach in mitigating the step length confounding problem.

%% file: S_Introduction.tex
\section{Introduction}

Recently, a variety of large reasoning models, \eg DeepSeek-R1 \citep{guo2025deepseek}, have achieved remarkable performance on complex reasoning tasks that demand long Chain-of-Thought (CoT) capabilities \citep{yang2025qwen3,qwen2025qwq32b}. To elicit such long CoT reasoning abilities in foundation models, Supervised Fine‑Tuning (SFT) on large‑scale, high‑quality datasets has become a standard paradigm \citep{chen2025acereason,guha2025openthoughts,zhao20251,yuan2026differential}.
Existing approaches typically begin by collecting complex mathematical and scientific problems, and then prompting stronger Large Language Models (LLMs) to generate answers as SFT datasets \citep{guha2025openthoughts,yuan2025naturalreasoning,huang2025loong}. Despite this pipeline effectively scaling up SFT data, such datasets still contain noisy instances, \eg incorrect reasoning steps \citep{zheng2025a} or overly complex reasoning trajectories \citep{sui2025stop}. To address this issue and build higher‑quality data subsets, LLM reasoning data selection has emerged as an active research topic \citep{ye2025limo,muennighoff2025s1}.

Generally, existing reasoning data selection methods often rely on heuristic rules, \eg verifiable answer correctness \citep{zhao20251,wu2025beyond}, response diversity \citep{jung2025prismatic,li2025exploring}, and problem difficulty \citep{muennighoff2025s1,guha2025openthoughts}.
These methods often depend heavily on manually crafted heuristics and do not consider the trained LLM’s adaptability to the SFT data. To overcome this limitation, the community introduces a \textbf{naturalness-based data selection} strategy \citep{zhang2025the,just2025distilling}, which involves computing the log probability assigned by an LLM to each SFT data sample and selecting those with higher average probabilities, as they are presumed to be better aligned with the LLM’s inherent preferences.

Unfortunately, our findings reveal that, when applied to long CoT datasets, \textit{the naturalness-based selection methods significantly prefer samples with longer reasoning steps} (\ie more tokens per step) rather than higher-adaptability ones. We refer to this phenomenon as the \textbf{step length confounding} problem in this work. We show in Fig.~\ref{mix_teacher_bias}, the step-length distribution of the \textit{selected} SFT data differs markedly from that of the \textit{unselected} data.
To further investigate the cause of this confounder, we build upon the quantitative results presented in Figs.~\ref{step_length_logits} and \ref{case_bias}. We observe that longer reasoning steps generally yield higher average log probabilities. This phenomenon can be explained by prior work showing that \textit{the first token of each reasoning step} often folks into different reasoning branches, thereby exhibiting higher entropy and consequently \textit{lower log probabilities} \citep{wang2025beyond,cheng2025reasoning}. \textit{Longer steps, however, dilute the impact of these low-probability first tokens, leading to a higher overall step log probability}, which in turn makes such longer-step examples more likely to be selected.

Given the above conclusion that low-probability first tokens lead to the step length confounding problem, we propose a mitigation method, namely \textbf{A}lleviating \textbf{S}tep \textbf{Le}ngth \textbf{C}onfounding (\textbf{\textsc{Aslec}}), which includes two variant approaches \textbf{\babya} and \textbf{\babyb}.
Specifically, \babya attempts to mitigate the confounding problem by simply dropping the first-token probabilities when computing the global average log probability. Despite this straightforward approach offering a preliminary mitigation to the confounding issue, it also entirely discards the contribution of the first token to data selection. Accordingly, \babyb, inspired by causal debiasing techniques \citep{udomcharoenchaikit2022mitigating}, fits a linear regression model to disentangle the first-token ratio as a confounding factor, and removes its effect when computing the global average log probability.

\def\huggingface{\raisebox{-1.5pt}{\includegraphics[height=1.0em]{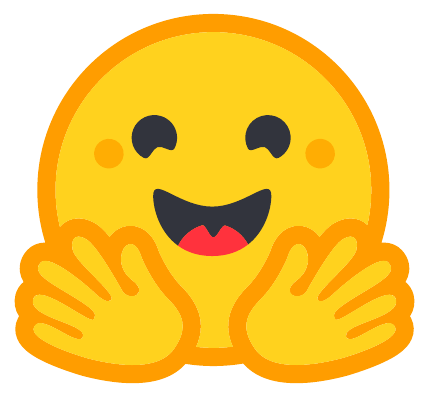}}}
\def\github{\raisebox{-1.5pt}{\includegraphics[height=1.0em]{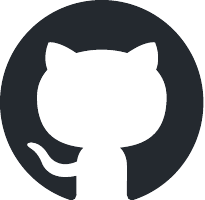}}}

In our experiments, we train four LLMs of varying sizes on two LLM reasoning benchmark datasets \textit{LIMO-v2} \citep{ye2025limo} and \textit{AceReason-1.1-SFT} \citep{chen2025acereason}, and evaluate the performance of different data selection strategies across five evaluation benchmarks. Our results demonstrate that the two proposed variants, \babya and \babyb, consistently outperform the state-of-the-art naturalness-based selection method, Local LP \citep{just2025distilling}, achieving average accuracy improvements of approximately 6.28\% and 9.08\%, respectively, across all model sizes and datasets.
\textit{Our source code and data are released in \github\ \url{https://github.com/wangbing1416/ASLEC}}. Meanwhile, in our implementation, we sample a large number of multi-source, multi-solution responses for \textit{LIMO-v2} and 10k \textit{AceReason-1.1-SFT} problems (average 64 responses per question). \textit{These large-scale SFT datasets are also be released in \huggingface\ \url{https://huggingface.co/collections/wangbing1416/msms-cot-sft}}.

Generally, our contributions can be summarized as the following three-fold:
\begin{itemize}
    \item Through extensive experiments, we identify a step length confounding problem in existing naturalness-based LLM reasoning data selection methods, and reveal that the cause lies in the low-probability first token of each step.
    \item We propose two variant methods, \babya and \babyb, which alleviate the step length confounding problem by intervening on the first-token probability when computing the global average log probability.
    \item Extensive experiments demonstrate the effectiveness of our proposed method and its ability to mitigate step length confounding.
\end{itemize}

%% file: S_Preliminary.tex
\section{Preliminary Experimental Analysis on Step Length Confounding} \label{sec.preliminary}

In this section, our preliminary experiments reveal that existing naturalness-based approaches for LLM reasoning data selection consistently suffer from \textbf{step length confounding}: they tend to prefer samples with longer reasoning steps.

\subsection{Naturalness-Based Data Selection}

Typically, an LLM reasoning SFT dataset is defined as $\mathcal{D} = \{\mathbf{q}_i, \mathbf{c}_i, \mathbf{a}_i\}_{i=1}^N$, where $\mathbf{q}$ denotes one question, and $\mathbf{c}$ and $\mathbf{a}$ represent its long CoT reasoning trajectory and final answer, respectively.  
The SFT objective is to optimize model parameters $\boldsymbol{\theta}$ by minimizing the negative log-likelihood of the target sequence $\mathbf{o}_i = \texttt{<think>}\ \mathbf{c}_i \ \texttt{</think>}\ \mathbf{a}_i$ as:
\begin{equation}
    \mathcal{L}_{\mathrm{SFT}}(\boldsymbol{\theta}) = - \frac{1}{N} \sum_{i=1}^N \sum_{t=1}^{|\mathbf{o}_i|} \log P_{\boldsymbol{\theta}} \left(o_{i,t} \mid o_{i,< t}, \mathbf{q}_i\right), \nonumber
\end{equation}
which is equal to the causal LM cross-entropy loss. While SFT typically treats all samples equally, data quality critically influences reasoning performance, as noisy or inconsistent trajectories can mislead learning. This motivates data selection strategies that prefer high-quality and informative subsets $\mathcal{\widehat D} \in \mathcal{D}$ to improve robustness. Among these works, \textbf{naturalness‑based methods} leverage the log probabilities produced by the LLM during SFT to select the data to which the model is best adapted. Formally, three representative methods are as follows:

\begin{itemize}
    \item \textbf{Log probabilities} \citep{zhang2025the} or \textbf{Perplexity} \citep{muennighoff2023scaling,yin2025compute} computes the geometric mean of the probabilities assigned to the target sequence outputs, as follows:
    \begin{align}
        s_i^{\mathrm{logp}} & = \frac{1}{|\mathbf{o}_i|} \sum_{t=1}^{|\mathbf{o}_i|} \log P_{\boldsymbol{\theta}}\left(o_{i,t} \mid \mathbf{o}_{i,<t}, \mathbf{q}_i\right), \nonumber \\
        s_i^{\mathrm{ppl}} & = \exp \left( - s_i^{\mathrm{logp}} \right).
    \end{align}
    A higher $s_i^{\mathrm{logp}}$ indicates that the model naturally adapts better to the given data.

    \item \textbf{Local log probabilities} \citep{just2025distilling} split the sequence $\mathbf{o}_i$ into steps $\mathcal{S}_i = \{\mathbf{s}_{ij}\}_{j=1}^{|\mathcal{S}_i|}$ by the token \texttt{\textbackslash n\textbackslash n} or sentences. For each step, it considers the question and its previous $k$ steps as context and calculates the geometric mean of its log probability accordingly.
    \begin{align}
        s_i^{\mathrm{loc}} & = \frac{1}{|\mathcal{S}_i|} \sum_{\mathbf{s}_{ij} \in \mathcal{S}_i} \frac{1}{|\mathbf{s}_{ij}|} \sum_{l=1}^{|\mathbf{s}_{ij}|} \log \\
        & P_{\boldsymbol{\theta}}\left(s_{ijl} \mid \mathbf{s}_{ij,<l}, \mathbf{s}_{i,\max(1,j-k):j-1}, \mathbf{q}_i\right). \nonumber
    \end{align}
    \item \textbf{Entropy} \citep{cui2025the,wang2025beyond} measures the average token-level uncertainty of the model's predictions. 
    \begin{align}
        s_i^{\mathrm{etp}} = \frac{1}{|\mathbf{o}_i|} \sum_{t=1}^{|\mathbf{o}_i|} \Big[ - \sum \nolimits _{v \in \mathcal{V}} & P_{\boldsymbol{\theta}}(v \mid o_{i,<t}, \mathbf{q}_i) \nonumber \\ 
        \log P_{\boldsymbol{\theta}} (v \mid & o_{i,<t}, \mathbf{q}_i) \Big],
    \end{align}
    where $\mathcal{V}$ represents the vocabulary, and lower entropy means the model is more confident in its outputs on the given example.
\end{itemize}
Existing naturalness‑based methods typically select a subset $\mathcal{\widehat D}$ from the large‑scale dataset $\mathcal{D}$ by either highest $s_i^{\mathrm{logp}}$ and $s_i^{\mathrm{loc}}$, or lowest $s_i^{\mathrm{ppl}}$ and $s_i^{\mathrm{etp}}$.

\subsection{Experimental Setup} \label{sec.2.2}

\noindent
\textbf{Models.} 
Our experiments utilize four long CoT reasoning LLMs of different families and parameters, including \textit{QwQ-32B} \citep{qwen2025qwq32b}, \textit{Qwen3-32B} \citep{yang2025qwen3}, \textit{DeepSeek-R1-Distill-Qwen-32B} \citep{guo2025deepseek}, and \textit{gpt-oss-120b} \citep{agarwal2025gpt}, as data sources for generating reasoning SFT data. We then use \textit{Qwen3-4B-Base} \citep{yang2025qwen3} as the target LLM to evaluate its log probabilities \wrt the SFT data. Detailed model cards for all LLMs are provided in Appendix~\ref{sec:appendix.settings.model}.

\vspace{2pt}
\noindent
\textbf{Benchmark and evaluation.}
We conduct our experiments on  \textit{LIMO-v2} \citep{ye2025limo}, a prevalent LLM reasoning benchmark comprising 800 carefully curated mathematical reasoning problems. For each problem, we generate 5 different responses from each of the 4 LLMs described above, using temperature sampling with $\tau = 0.6$. From the generated $4 \times 5 = 20$ responses per problem, we select 5 final responses using one of the four representative naturalness‑based data selection methods, (i) \textit{GRACE} \citep{zhang2025the}, which selects the responses with the highest $s^{\mathrm{logp}}$; (ii) \textit{Local LP} \citep{just2025distilling}, which also selects the responses with the highest $s^{\mathrm{loc}}$; (iii) \textit{Min Entropy} \citep{cui2025the}, which selects the responses with the smallest $s^{\mathrm{etp}}$; (iv) \textit{Min Perplex}, which select the responses with the smallest $s_i^{\mathrm{ppl}}$, to analyze the step length confounding phenomenon. More details on the benchmarks and experimental setup are provided in Appendix~\ref{sec:appendix.settings.data}, and additional analysis across more datasets, \eg \textit{AceReason-1.1-SFT}, and more LLMs can be found in Appendix~\ref{sec:appendix.analysis}.

\subsection{Results on Step Length Confounding}

Through the preliminary experiments in this section, we find that these naturalness-based methods suffer from the step length confounding issue.

\begin{figure}[t]
  \centering
  \includegraphics[width=1.0\columnwidth]{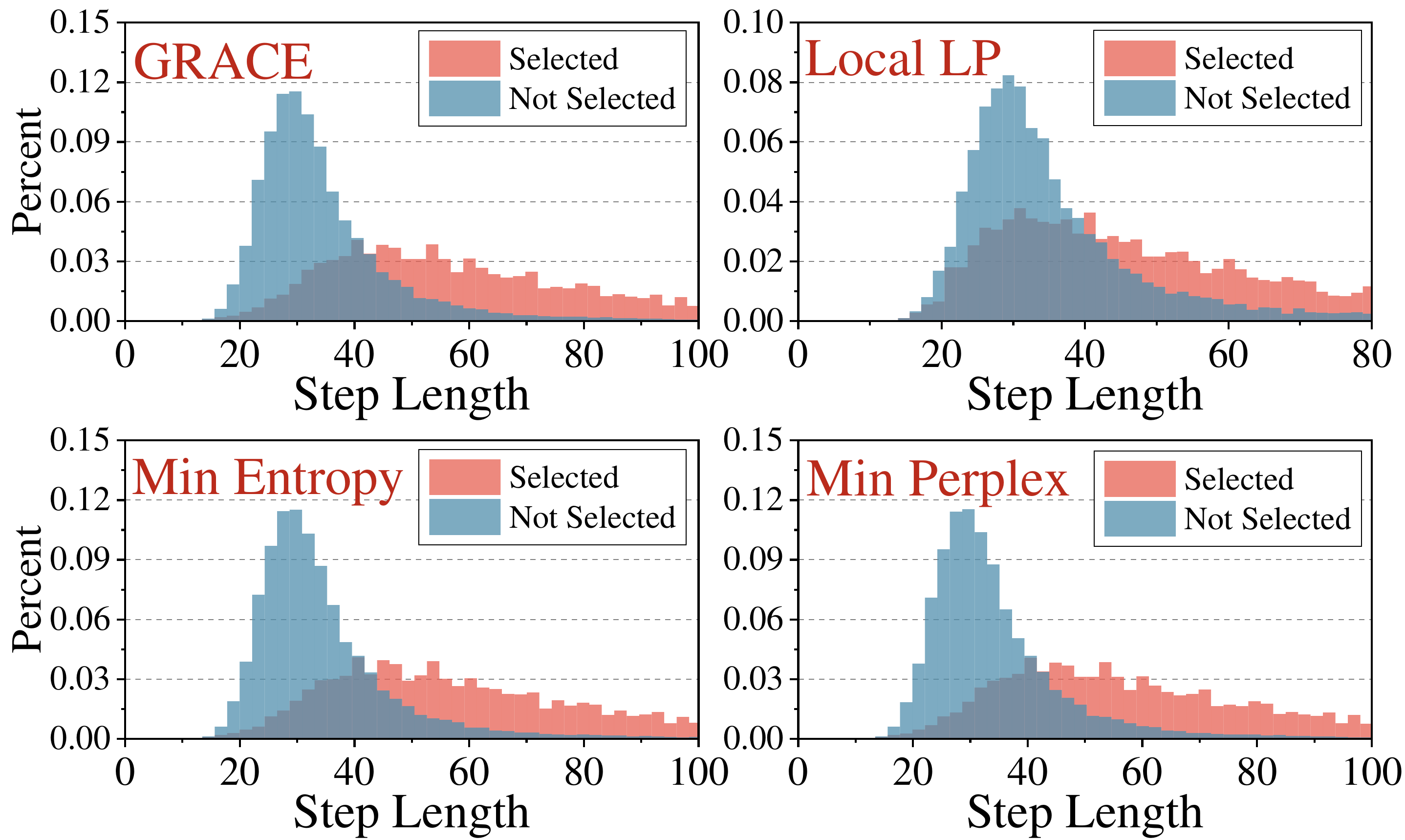}
  \footnotesize
  \vspace{-0.5cm}
  \caption{Step length distribution of data samples \textit{selected} and \textit{unselected} by different naturalness-based data selection methods.}
  \label{mix_teacher_bias}
\end{figure}

\vspace{2pt} \noindent
\textbf{Results and analysis.}
In Fig.~\ref{mix_teacher_bias}, we illustrate the selection difference of naturalness‑based methods across the 16,000 responses (800 problems × 20 responses each) generated by the four LLMs. 
Fig.~\ref{mix_teacher_bias} compares the step-length distributions of responses \textit{selected} versus \textit{not selected} by four naturalness-based data selection methods. Across all methods, selected responses consistently exhibit longer step lengths, whereas the step lengths of unselected responses are more concentrated at lower values, with an average of approximately 30. This pattern underscores the consistent influence of step length on the decisions made by these naturalness-based criteria. \footnote{Notably, data selection actually also correlates with total response length (i.e., avg. $L$). However, as discussed in Appendix~\ref{sec:app.total}, \textbf{the effect of total response length is substantially weaker than that of step length}.}
Based on these observations, we formulate the following conclusion and refer to this phenomenon as \textit{step length confounding}.

\begin{tcolorbox}[
  colframe=black,
  boxrule=0.8pt,
  left=2.5pt, right=2.5pt, top=2.5pt, bottom=2.5pt, 
  fonttitle=\bfseries
]
\linespread{0.95}\selectfont
\textbf{$\boldsymbol{\star}$ Conclusion 1.}
The naturalness-based reasoning data selection approach tends to \textbf{prefer samples with longer reasoning steps} (\ie more tokens per step).
\end{tcolorbox}

\begin{figure}[t]
  \centering
  \includegraphics[width=1.0\columnwidth]{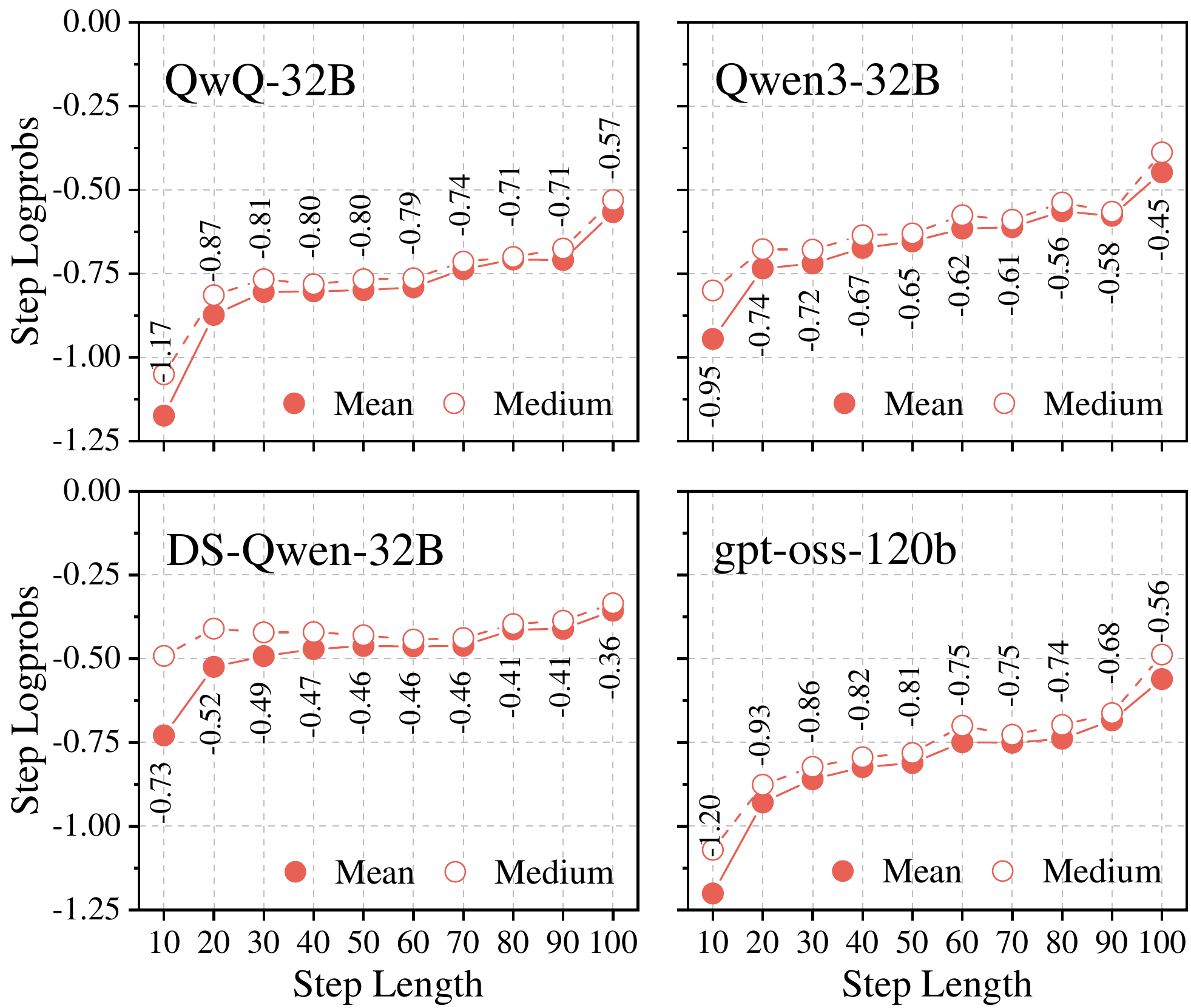}
  \caption{Relationship between step‑level log probability and step length.}
  \label{step_length_logits}
  \vspace{-3pt}
\end{figure}

\subsection{Why Step Length Confounding?} \label{sec:whybias}

Given the step length confounding problem in LLM reasoning data selection, we seek to figure out the intrinsic causes resulting in this issue. Therefore, we give the following further empirical evidence.

\vspace{2pt} \noindent
\textbf{For longer steps, the model assigns higher step-level log probabilities.}
We first investigate the relationship between step length and the average log probability per step. As illustrated in Fig.~\ref{step_length_logits}, outputs from different LLMs are segmented into steps. For steps of different lengths, we compute the average step‑level log probabilities assigned by the target LLM \textit{Qwen3-4B-Base}. The results reveal a clear pattern: longer reasoning steps consistently receive higher step‑level log probabilities, and a monotonic increasing relationship is observed between step length and log probability.

\begin{figure}[t]
  \centering
  \includegraphics[width=1.0\columnwidth]{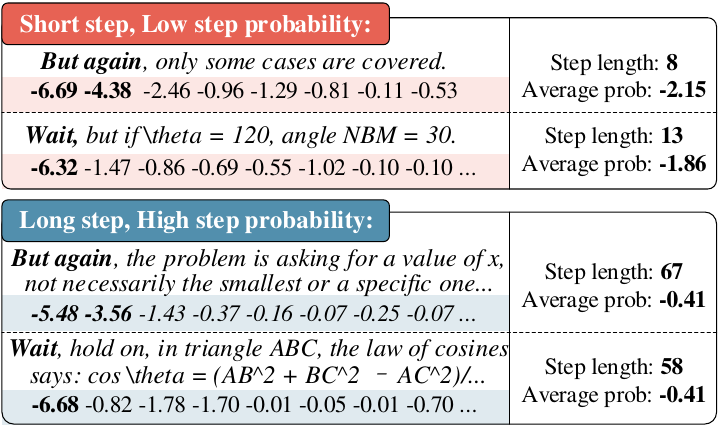}
  \caption{Representative cases illustrating token‑level log probabilities for varying step lengths.}
  \label{case_bias}
\end{figure}

\vspace{2pt} \noindent
\textbf{For longer steps, the ratio of low-probability first tokens is lower.}
To further investigate the cause of the monotonic relationship between step length and step-level log probability, we examine several representative examples in Fig.~\ref{case_bias}, which illustrate short steps with low log probabilities and long steps with high log probabilities, respectively.
Across all steps, \textit{the first token consistently exhibits a lower log probability}. Previous studies have also confirmed this phenomenon \citep{wang2025beyond,cheng2025reasoning}, attributing it to the fact that minority first tokens at each step often fork branches toward diverse reasoning pathways. Such branching behavior introduces higher entropy, which in turn yields lower log probabilities.
Therefore, in longer steps, such low-probability first tokens always constitute a smaller proportion of the total tokens. Consequently, the larger number of non‑first tokens dilutes the lower log probabilities of these first tokens, leading to a higher overall log probability and making such samples more likely to be selected by naturalness‑based methods.
In summary, our experiments lead to the following conclusion:
\begin{tcolorbox}[
  colframe=black,
  boxrule=0.8pt,
  left=2.5pt, right=2.5pt, top=2.5pt, bottom=2.5pt, 
  fonttitle=\bfseries
]
\linespread{0.95}\selectfont
\textbf{$\boldsymbol{\star}$ Conclusion 2.}
In naturalness‑based methods, step length confounding occurs because \textbf{the low-probability first token constitutes a smaller ratio of longer responses}. This increases the average log probabilities, making samples with longer steps more likely to be selected.
\end{tcolorbox}
\vspace{-2pt}
Based on the above observations, we seek to design a debiasing approach targeting the first token to address the step length confounding problem.

%% file: S_Method.tex
\section{The Proposed Method}

In this section, we present our proposed variant methods \babya and \babyb for LLM reasoning SFT data selection in detail.

\noindent
\textbf{Problem definition.}
Given $i$-th complex question $\mathbf{q}_i$ in the LLM reasoning SFT dataset $\mathcal{D}$, and $K$ different responses $\mathbf{o}_i^k = \texttt{<think>}\ \mathbf{c}_i^k \ \texttt{</think>}\ \mathbf{a}_i^k$, $k \in \{1, \cdots, K\}$, where each $\mathbf{c}_i^k$ represents a reasoning trajectory and $\mathbf{a}_i^k$ the corresponding answer. These multiple responses may vary not only in correctness but also in reasoning quality, verbosity, and step length. Our method defines two metrics $s_i^{\mathrm{drop}}$ and $s_i^{\mathrm{casl}}$ to select one or more responses that best align with the trained reasoning LLM and are not confounded by the step length.

\subsection{\babya: Dropping the First Token}

As analyzed in Sec.~\ref{sec:whybias}, we attribute this step length confounding problem to the influence of the first token’s probabilities \citep{bu2025beyond}. Consequently, the most straightforward approach is to drop the first token at each step when computing the geometric mean of the probabilities. Formally, we split a solution $\mathbf{o}_i$ into $L$ reasoning steps $\mathcal{S}_i = \{\mathbf{s}_i^l\}_{l=1}^L$ and compute the metric as follows:
\begin{equation}
    \label{eq4}
    \begin{aligned}
    s_i^{\mathrm{drop}} = \frac{1}{|\mathbf{o}_i| - |\mathcal{S}_i|} & \sum \nolimits_{\mathbf{s}_i^l \in \mathcal{S}_i} \sum \nolimits_{t=2}^{|\mathbf{s}_i^l|} \\
    \log & P_{\boldsymbol{\theta}} \left(s_{i,t}^l \mid \mathbf{s}_{i,<t}^l, \mathbf{s}_{i}^{<l}, \mathbf{q}_i\right),
    \end{aligned}
\end{equation}
where $\mathbf{s}_{i,<t}^l$ denotes the first $t$ tokens of each step, and $\mathbf{s}_{i}^{<l}$ denotes all tokens across the first $l$ steps.

\subsection{\babyb: Causally De-biasing} \label{sec.3.2}

Although dropping the first token mitigates the bias it introduces, it simultaneously discards potentially informative signals carried by the first token itself. To address this trade-off, we draw inspiration from causal debiasing methods \citep{udomcharoenchaikit2022mitigating,zhu2022generalizing}, treating step length as a confounding factor and applying appropriate adjustments to account for its influence.
To formalize this intuition, the log probability $s_i^{\mathrm{logp}}$ can be decomposed as the following linear regression equation:
\begin{equation}
    \label{eq5}
    s_i^{\mathrm{logp}} = \beta_1 s_i^{\mathrm{first}} + \beta_2 s_i^{\mathrm{drop}} + \gamma \mathcal{Z}_i + \epsilon, 
\end{equation}
where $s_i^{\mathrm{first}}$ and $s_i^{\mathrm{drop}}$ represent the average log probabilities of the first token and the tokens excluding the first one in Eq.~\eqref{eq4}, respectively; $\mathcal{Z}_i$ represents the confounding factor, which in our method is defined as the proportion of the first token among all tokens; and $\epsilon$ denotes a residual noise term. The notation $s_i^{\mathrm{first}}$ and $\mathcal{Z}_i$ are formally given by:
\begin{equation}
    \label{eq6}
    \begin{gathered}
        s_i^{\mathrm{first}} = \frac{1}{|\mathcal{S}_i|} \sum _{\mathbf{s}_i^l \in \mathcal{S}_i} \log P_{\boldsymbol{\theta}}\left(s_{i,1}^l \mid \mathbf{s}_{i}^{<l}, \mathbf{q}_i\right), \\
        \mathcal{Z}_i = \frac{|\mathcal{S}_i|}{|\mathbf{o}_i|},
    \end{gathered}
\end{equation}
where $s_{i,1}^l$ denotes the first token in the step $\mathbf{s}_{i}^{l}$. 
The basic idea of \babyb is to adjust the raw log-probability score by removing the estimated influence of the confounder $\mathcal{Z}_i$. This yields a deconfounded metric $s_i^{\mathrm{casl}}$, defined as:
\begin{equation}
    s_i^{\mathrm{casl}} \sim s_i^{\mathrm{logp}} - \gamma \mathcal{Z}_i.
\end{equation}
Accordingly, to calculate $s_i^{\mathrm{casl}}$ by estimating $\gamma$, given the dataset $\{s_i^{\mathrm{logp}}, s_i^{\mathrm{first}}, s_i^{\mathrm{drop}}, \mathcal{Z}_i\}_{i=1}^N$, the parameters $\beta_1$, $\beta_2$, $\gamma$ are estimated via ordinary least squares:
\begin{equation}
    \label{eq7}
    \mathop{\boldsymbol{\min}} \limits_{\beta_1,\beta_2,\gamma} \sum_{i=1}^N \left( s_i^{\mathrm{logp}}-\beta_1 s_i^{\mathrm{first}} - \beta_2 s_i^{\mathrm{drop}} - \gamma \mathcal{Z}_i \right)^2.
\end{equation}

This optimization admits a closed-form solution. Then the parameter vector is obtained as follows:
\begin{equation}
    \label{eq8}
    \left[ \beta_1,\beta_2,\gamma \right]^\top = \big(\mathbf{X}^\top \mathbf{X}\big)^{-1} \mathbf{X}^\top \mathbf{Y},
\end{equation}
\begin{equation}
    \label{eq9}
    \mathbf{X}_{i,:} = \left[ s_i^{\mathrm{first}}, s_i^{\mathrm{drop}}, \mathcal{Z}_i \right], 
    \ \mathbf{Y}_{i,:} = \left[ s_i^{\mathrm{logp}} \right].
\end{equation}
Once $\gamma$ is estimated, we compute the final deconfounded score $s_i^{\mathrm{casl}} = s_i^{\mathrm{logp}} - \gamma \mathcal{Z}_i$ for each instance and use it for downstream data selection.

%% file: S_Experiment.tex
\section{Experimental Evaluation}

\begin{table*}[t]
\centering
\renewcommand\arraystretch{1.05}
  \caption{Experimental results on \textit{LIMO-v2} \citep{ye2025limo}. We generate five responses per source LLM, and select five responses from these ones (select 4k responses from 16k data). The bold results represent the best scores.}
  \label{limoresult}
  \small
  \setlength{\tabcolsep}{5pt}{
  \begin{tabular}{m{0.25cm}<{\centering}m{3.30cm}m{1.38cm}m{1.38cm}m{1.38cm}m{1.38cm}m{1.45cm}m{1.45cm}m{0.85cm}<{\centering}}
    \toprule
    \multirow{2}{*}{\rotatebox{90}{\textbf{Qwen}}} & \multirow{2}{*}{\quad \quad \quad \quad Method} & \multicolumn{2}{c}{\textit{AIME24}} & \multicolumn{2}{c}{\textit{AIME25}} & \multirow{2}{*}{\textit{MATH500}} & \multirow{2}{*}{\textit{OlympicB}} & \multirow{2}{*}{\textbf{Avg.}} \\
    \cmidrule(r){3-4} \cmidrule(r){5-6}
    & & Accuracy & Pass@4 & Accuracy & Pass@4 &  &  &  \\
    \hline
    \multirow{4}{*}{\rotatebox{90}{\textbf{4B-Base} }}
    & + GRACE {\scriptsize \citep{zhang2025the}}
      & 16.66 & 30.00 & 15.83 & 33.33 & 59.40 & 33.33 & 31.42 \\
    & + Local LP {\scriptsize \citep{just2025distilling}}
      & 19.16 & 36.66 & 20.83 & 36.66 & 71.60 & 34.11 & 36.50 \\
    & \cellcolor{lightgrayv}+ \babya {\scriptsize\ (ours)} 
      & \cellcolor{lightgrayv}30.00{\color{myred} \scriptsize $\uparrow$10.84} 
      & \cellcolor{lightgrayv}50.00{\color{myred} \scriptsize $\uparrow$13.34} 
      & \cellcolor{lightgrayv}28.33{\color{myred} \scriptsize $\uparrow$7.50} 
      & \cellcolor{lightgrayv}43.33{\color{myred} \scriptsize $\uparrow$6.67} 
      & \cellcolor{lightgrayv}77.80{\color{myred} \scriptsize $\uparrow$6.20} 
      & \cellcolor{lightgrayv}38.38{\color{myred} \scriptsize $\uparrow$4.27} 
      & \cellcolor{lightgrayv}44.64 \\
    & \cellcolor{lightgrayv}+ \babyb {\scriptsize\ (ours)} 
      & \cellcolor{lightgrayv}\textbf{31.66}{\color{myred} \scriptsize $\uparrow$12.50} 
      & \cellcolor{lightgrayv}\textbf{53.33}{\color{myred} \scriptsize $\uparrow$16.67} 
      & \cellcolor{lightgrayv}\textbf{30.83}{\color{myred} \scriptsize $\uparrow$10.00} 
      & \cellcolor{lightgrayv}\textbf{46.66}{\color{myred} \scriptsize $\uparrow$10.00} 
      & \cellcolor{lightgrayv}\textbf{80.00}{\color{myred} \scriptsize $\uparrow$8.40} 
      & \cellcolor{lightgrayv}\textbf{42.81}{\color{myred} \scriptsize $\uparrow$8.70} 
      & \cellcolor{lightgrayv}\textbf{47.54} \\
    \hline
    \multirow{4}{*}{\rotatebox{90}{\textbf{8B-Base}}}
    & + GRACE {\scriptsize \citep{zhang2025the}}
      & 30.83 & 53.33 & 21.66 & 36.66 & 72.00 & 39.70 & 42.36 \\
    & + Local LP {\scriptsize \citep{just2025distilling}}
      & 34.16 & 53.33 & 20.83 & 36.66 & 76.60 & 42.81 & 44.06 \\
    & \cellcolor{lightgrayv}+ \babya {\scriptsize\ (ours)} 
      & \cellcolor{lightgrayv}41.66{\color{myred} \scriptsize $\uparrow$10.50} 
      & \cellcolor{lightgrayv}\textbf{66.66}{\color{myred} \scriptsize $\uparrow$13.33} 
      & \cellcolor{lightgrayv}36.66{\color{myred} \scriptsize $\uparrow$15.83} 
      & \cellcolor{lightgrayv}43.33{\color{myred} \scriptsize $\uparrow$6.67} 
      & \cellcolor{lightgrayv}81.40{\color{myred} \scriptsize $\uparrow$4.80} 
      & \cellcolor{lightgrayv}47.85{\color{myred} \scriptsize $\uparrow$5.04} 
      & \cellcolor{lightgrayv}52.92 \\
    & \cellcolor{lightgrayv}+ \babyb {\scriptsize\ (ours)}
      & \cellcolor{lightgrayv}\textbf{45.00}{\color{myred} \scriptsize $\uparrow$13.34} 
      & \cellcolor{lightgrayv}\textbf{66.66}{\color{myred} \scriptsize $\uparrow$13.33} 
      & \cellcolor{lightgrayv}\textbf{37.50}{\color{myred} \scriptsize $\uparrow$16.67} 
      & \cellcolor{lightgrayv}\textbf{53.33}{\color{myred} \scriptsize $\uparrow$16.67} 
      & \cellcolor{lightgrayv}\textbf{85.40}{\color{myred} \scriptsize $\uparrow$8.80} 
      & \cellcolor{lightgrayv}\textbf{49.03}{\color{myred} \scriptsize $\uparrow$6.22} 
      & \cellcolor{lightgrayv}\textbf{56.15} \\
    \hline
    \multirow{4}{*}{\rotatebox{90}{\fontsize{8pt}{10pt} \textbf{4B-Instruct}\ }}
    & + GRACE {\scriptsize \citep{zhang2025the}}
      & 59.16 & 73.33 & 50.00 & 73.33 & 79.36 & 47.79 & 63.82 \\
    & + Local LP {\scriptsize \citep{just2025distilling}}
      & 61.66 & 80.00 & 49.16 & 73.33 & 80.75 & 50.14 & 65.84 \\
    & \cellcolor{lightgrayv}+ \babya {\scriptsize\ (ours)}
      & \cellcolor{lightgrayv}69.16{\color{myred} \scriptsize $\uparrow$7.50}
      & \cellcolor{lightgrayv}83.33{\color{myred} \scriptsize $\uparrow$3.33}
      & \cellcolor{lightgrayv}56.66{\color{myred} \scriptsize $\uparrow$7.50}
      & \cellcolor{lightgrayv}80.00{\color{myred} \scriptsize $\uparrow$6.67}
      & \cellcolor{lightgrayv}89.88{\color{myred} \scriptsize $\uparrow$9.13}
      & \cellcolor{lightgrayv}57.64{\color{myred} \scriptsize $\uparrow$7.50}
      & \cellcolor{lightgrayv}72.77 \\
    & \cellcolor{lightgrayv}+ \babyb {\scriptsize\ (ours)}
      & \cellcolor{lightgrayv}\textbf{71.66}{\color{myred} \scriptsize $\uparrow$10.00}
      & \cellcolor{lightgrayv}\textbf{90.00}{\color{myred} \scriptsize $\uparrow$10.00}
      & \cellcolor{lightgrayv}\textbf{58.33}{\color{myred} \scriptsize $\uparrow$9.17}
      & \cellcolor{lightgrayv}\textbf{83.33}{\color{myred} \scriptsize $\uparrow$10.00}
      & \cellcolor{lightgrayv}\textbf{93.20}{\color{myred} \scriptsize $\uparrow$12.45}
      & \cellcolor{lightgrayv}\textbf{60.44}{\color{myred} \scriptsize $\uparrow$10.30}
      & \cellcolor{lightgrayv}\textbf{76.16} \\
    \hline
    \multirow{4}{*}{\rotatebox{90}{\fontsize{8pt}{10pt} \textbf{7B-Instruct}\ }}
    & + GRACE {\scriptsize \citep{zhang2025the}}
      & 17.50 & 26.66 & 11.66 & 23.33 & 61.50 & 32.35& 28.83 \\
    & + Local LP {\scriptsize \citep{just2025distilling}}
      & 17.50 & 30.00 & 10.83 & 26.66 & 64.68 & 33.97 & 30.60 \\
    & \cellcolor{lightgrayv}+ \babya {\scriptsize\ (ours)}
      & \cellcolor{lightgrayv}24.16{\color{myred} \scriptsize $\uparrow$6.66}
      & \cellcolor{lightgrayv}40.00{\color{myred} \scriptsize $\uparrow$10.00}
      & \cellcolor{lightgrayv}20.83{\color{myred} \scriptsize $\uparrow$10.00}
      & \cellcolor{lightgrayv}36.66{\color{myred} \scriptsize $\uparrow$10.00}
      & \cellcolor{lightgrayv}80.60{\color{myred} \scriptsize $\uparrow$15.92}
      & \cellcolor{lightgrayv}41.17{\color{myred} \scriptsize $\uparrow$7.20}
      & \cellcolor{lightgrayv}40.57 \\
    & \cellcolor{lightgrayv}+ \babyb {\scriptsize\ (ours)}
      & \cellcolor{lightgrayv}\textbf{28.33}{\color{myred} \scriptsize $\uparrow$10.83}
      & \cellcolor{lightgrayv}\textbf{46.66}{\color{myred} \scriptsize $\uparrow$16.66}
      & \cellcolor{lightgrayv}\textbf{24.16}{\color{myred} \scriptsize $\uparrow$13.33}
      & \cellcolor{lightgrayv}\textbf{46.66}{\color{myred} \scriptsize $\uparrow$20.00}
      & \cellcolor{lightgrayv}\textbf{81.60}{\color{myred} \scriptsize $\uparrow$16.92}
      & \cellcolor{lightgrayv}\textbf{45.18}{\color{myred} \scriptsize $\uparrow$11.21}
      & \cellcolor{lightgrayv}\textbf{45.43} \\
    \bottomrule
  \end{tabular} }
\end{table*}

In this section, we empirically evaluate the performance of our two proposed variant methods. 

\vspace{2pt} \noindent
\textbf{Evaluation settings.}
The experiments are conducted on two datasets, \textit{LIMO-v2} and \textit{AceReason-1.1-SFT}, using four different families of source LLMs:  \textit{QwQ-32B}, \textit{Qwen3-32B}, \textit{DeepSeek-R1-Distill-Qwen-32B}, and \textit{gpt-oss-120b}, and four target LLMs of varying sizes: \textit{Qwen3-4B-Base}, \textit{Qwen3-8B-Base}, \textit{Qwen3-4B-Instruct}, and \textit{Qwen2.5-7B-Instruct}. Detailed descriptions of these LLMs and the implementation details of our SFT training are provided in Appendix~\ref{sec:appendix.settings}.
We evaluate our trained LLMs on five benchmarks, including four mathematical reasoning datasets, \textit{AIME24}, \textit{AIME25}, \textit{MATH500} \citep{lightman2023let}, and \textit{OlympiadBench} \citep{he2024olympiadbench}, as well as one challenging scientific reasoning dataset \textit{GPQA} \citep{rein2024gpqa}. 
In addition, we compare two naturalness-based data selection methods: (i) GRAPE: selecting the responses with the highest $s^{\mathrm{logp}}$; and (ii) Local LP: selecting the responses with the highest $s^{\mathrm{loc}}$.

\subsection{Main Results}

Tables~\ref{limoresult} and \ref{acereasonresult} present the experimental results of the four target LLMs on the \textit{LIMO-v2} and \textit{AceReason-1.1-SFT} datasets, respectively. Overall, both variants of our approach outperform the existing naturalness‑based selection method, achieving average accuracy gains of 6.28\% and 9.08\% over the SOTA method Local LP \citep{just2025distilling} on the two datasets.
More specifically, prior naturalness‑based methods, \eg GRACE and Local LP, are often hindered by the step length confounding problem, which leads them to overly prefer samples from a single data source. Consequently, their training performance consistently degrades and falls significantly below our method, which samples more evenly across diverse sources.

When comparing the two variant methods, \babyb consistently outperforms \babya. This result suggests that the causal debiasing strategy successfully preserves the informative patterns embedded in the probability distribution of the first tokens.
Meanwhile, the results indicate that our debiasing strategies are particularly effective when data or model capacity is limited. For example, the performance gain on \textit{LIMO-v2} exceeds that on \textit{AceReason-1.1-SFT}, highlighting their strong suitability for low-resource SFT scenarios. In such settings, low-quality samples tend to have a more pronounced negative effect; our methods mitigate step length confounding and thereby enhance generalization performance.

We further compare our methods by training on \textit{LIMO‑v2} and evaluating on \textit{GPQA}, a benchmark on the scientific domain. The experimental results again show that our approaches consistently and significantly outperform the SOTA Local LP selection baseline, and that the \babyb variant achieves better performance than \babya.

\begin{table}[t]
\centering
\renewcommand\arraystretch{1.05}
  \caption{Experimental performance on \textit{GPQA}.}
  \label{gpqaresults}
  \small
  \setlength{\tabcolsep}{5pt}{
  \setlength{\aboverulesep}{1.2pt}
 \setlength{\belowrulesep}{1.2pt}
  \begin{tabular}{m{1.8cm}<{\centering}m{0.95cm}<{\centering}m{1.0cm}<{\centering}m{1.1cm}<{\centering}m{1.1cm}<{\centering}}
    \toprule
    \multirow{2}{*}{Method} & Acc. & Pass@4 & Acc. & Pass@4 \\
    \cmidrule(r){2-3} \cmidrule(r){4-5}
    & \multicolumn{2}{c}{\fontsize{8pt}{11pt}\selectfont \textbf{Qwen3-4B-Base}} & \multicolumn{2}{c}{\fontsize{8pt}{11pt}\selectfont \textbf{Qwen3-4B-Insruct}} \\

    \hline
    GRACE & 28.15 & 60.10 & 50.37 & 75.75 \\
    Local LP & 29.16 & 61.61 & 52.14 & 77.27 \\
    \rowcolor{lightgrayv} \babya & 34.97 & 65.65 & 58.83 & 83.33 \\
    \rowcolor{lightgrayv} \babyb & \textbf{35.35} & \textbf{66.66} & \textbf{61.23} & \textbf{84.34} \\
    \hline
    \multirow{2}{*}{Method} & Acc. & Pass@4 & Acc. & Pass@4 \\
    \cmidrule(r){2-3} \cmidrule(r){4-5}
    & \multicolumn{2}{c}{\fontsize{8pt}{11pt}\selectfont \textbf{Qwen3-8B-Base}} & \multicolumn{2}{c}{\fontsize{8pt}{11pt}\selectfont \textbf{Qwen2.5-7B-Insruct}} \\
    \hline
    GRACE & 47.97 & 75.25 & 25.37 & 56.06 \\
    Local LP & 49.49 & 77.77 & 26.13 & 57.57 \\
    \rowcolor{lightgrayv} \babya & 51.01 & 79.79 & 35.98 & 67.17 \\
    \rowcolor{lightgrayv} \babyb & \textbf{52.14} & \textbf{82.32} & \textbf{38.51} & \textbf{74.74} \\
    \bottomrule
  \end{tabular} }
\end{table}

\begin{figure}[t]
  \centering
  \includegraphics[width=1.0\columnwidth]{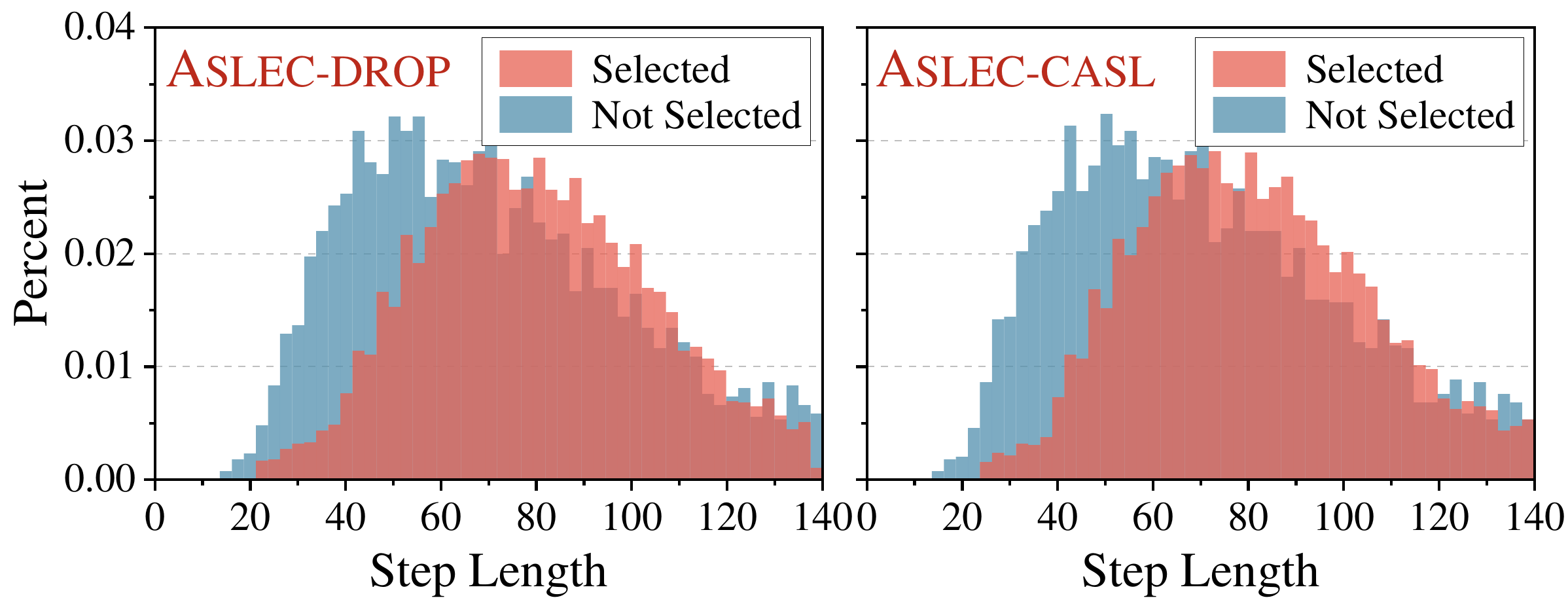}
  \caption{Step length distributions for data \textit{selected} versus \textit{unselected} by our two proposed variant methods.}
  \label{step_length_ours}
\end{figure}

\begin{table*}[t]
\centering
\renewcommand\arraystretch{1.08}
  \caption{Experimental results on \textit{AceReason-1.1-SFT} \citep{chen2025acereason}. We generate one response per source LLM, and select one response from these responses (select 10k responses from 40k data).}
  \label{acereasonresult}
  \small
  \setlength{\tabcolsep}{5pt}{
  \begin{tabular}{m{0.25cm}<{\centering}m{3.30cm}m{1.38cm}m{1.38cm}m{1.38cm}m{1.38cm}m{1.45cm}m{1.45cm}m{0.85cm}<{\centering}}
    \toprule
    \multirow{2}{*}{\rotatebox{90}{\textbf{Qwen}}} & \multirow{2}{*}{\quad \quad \quad \quad Method} & \multicolumn{2}{c}{\textit{AIME24}} & \multicolumn{2}{c}{\textit{AIME25}} & \multirow{2}{*}{\textit{MATH500}} & \multirow{2}{*}{\textit{OlympicB}} & \multirow{2}{*}{\textbf{Avg.}} \\
    \cmidrule(r){3-4} \cmidrule(r){5-6}
    & & Accuracy & Pass@4 & Accuracy & Pass@4 &  &  &  \\
    \hline
    % 4B-Base
    \multirow{4}{*}{\rotatebox{90}{\textbf{4B-Base} }}
    & + GRACE {\scriptsize \citep{zhang2025the}}
      & 30.00 & 60.00 & 29.16 & 36.66 & 77.20 & 42.50 & 43.82 \\
    & + Local LP {\scriptsize \citep{just2025distilling}}
      & 32.50 & 63.33 & 30.00 & 40.00 & 77.80 & 43.08 & 45.22 \\
    & \cellcolor{lightgrayv}+ \babya {\scriptsize\ (ours)} 
      & \cellcolor{lightgrayv}41.66{\color{myred} \scriptsize $\uparrow$9.16} 
      & \cellcolor{lightgrayv}\textbf{73.33}{\color{myred} \scriptsize $\uparrow$10.00} 
      & \cellcolor{lightgrayv}\textbf{30.83}{\color{myred} \scriptsize $\uparrow$0.83} 
      & \cellcolor{lightgrayv}43.33{\color{myred} \scriptsize $\uparrow$3.33} 
      & \cellcolor{lightgrayv}84.00{\color{myred} \scriptsize $\uparrow$6.20} 
      & \cellcolor{lightgrayv}47.20{\color{myred} \scriptsize $\uparrow$4.12} 
      & \cellcolor{lightgrayv}46.48 \\
    & \cellcolor{lightgrayv}+ \babyb {\scriptsize\ (ours)} 
      & \cellcolor{lightgrayv}\textbf{46.66}{\color{myred} \scriptsize $\uparrow$14.16} 
      & \cellcolor{lightgrayv}\textbf{73.33}{\color{myred} \scriptsize $\uparrow$10.00} 
      & \cellcolor{lightgrayv}\textbf{30.83}{\color{myred} \scriptsize $\uparrow$0.83} 
      & \cellcolor{lightgrayv}\textbf{46.66}{\color{myred} \scriptsize $\uparrow$6.66} 
      & \cellcolor{lightgrayv}\textbf{84.60}{\color{myred} \scriptsize $\uparrow$6.80} 
      & \cellcolor{lightgrayv}\textbf{48.23}{\color{myred} \scriptsize $\uparrow$5.15} 
      & \cellcolor{lightgrayv}\textbf{47.50} \\
    \hline
    % 8B-Base
    \multirow{4}{*}{\rotatebox{90}{\textbf{8B-Base}}}
    & + GRACE {\scriptsize \citep{zhang2025the}}
      & 40.83 & 66.66 & 29.16 & 43.33 & 75.00 & 44.11 & 49.46 \\
    & + Local LP {\scriptsize \citep{just2025distilling}}
      & 42.50 & 70.00 & 29.16 & 43.33 & 77.60 & 44.41 & 50.71 \\
    & \cellcolor{lightgrayv}+ \babya {\scriptsize\ (ours)} 
      & \cellcolor{lightgrayv}50.00{\color{myred} \scriptsize $\uparrow$7.50} 
      & \cellcolor{lightgrayv}\textbf{76.66}{\color{myred} \scriptsize $\uparrow$6.66} 
      & \cellcolor{lightgrayv}36.66{\color{myred} \scriptsize $\uparrow$7.50} 
      & \cellcolor{lightgrayv}\textbf{53.33}{\color{myred} \scriptsize $\uparrow$10.00} 
      & \cellcolor{lightgrayv}86.20{\color{myred} \scriptsize $\uparrow$8.60} 
      & \cellcolor{lightgrayv}49.33{\color{myred} \scriptsize $\uparrow$4.92} 
      & \cellcolor{lightgrayv}54.60 \\
    & \cellcolor{lightgrayv}+ \babyb {\scriptsize\ (ours)}
      & \cellcolor{lightgrayv}\textbf{53.33}{\color{myred} \scriptsize $\uparrow$10.83} 
      & \cellcolor{lightgrayv}\textbf{76.66}{\color{myred} \scriptsize $\uparrow$6.66} 
      & \cellcolor{lightgrayv}\textbf{39.16}{\color{myred} \scriptsize $\uparrow$10.00} 
      & \cellcolor{lightgrayv}\textbf{53.33}{\color{myred} \scriptsize $\uparrow$10.00} 
      & \cellcolor{lightgrayv}\textbf{86.60}{\color{myred} \scriptsize $\uparrow$9.00} 
      & \cellcolor{lightgrayv}\textbf{51.02}{\color{myred} \scriptsize $\uparrow$6.61} 
      & \cellcolor{lightgrayv}\textbf{56.21} \\
    \hline
    % 4B-Instruct
    \multirow{4}{*}{\rotatebox{90}{\fontsize{8pt}{10pt} \textbf{4B-Instruct}\ }}
    & + GRACE {\scriptsize \citep{zhang2025the}}
      & 54.16 & 73.33 & 43.33 & 63.33 & 84.80 & 48.82 & 59.65 \\
    & + Local LP {\scriptsize \citep{just2025distilling}}
      & 57.50 & 76.66 & 42.50 & 66.66 & 85.60 & 49.26 & 61.79 \\
    & \cellcolor{lightgrayv}+ \babya {\scriptsize\ (ours)}
      & \cellcolor{lightgrayv}64.16{\color{myred} \scriptsize $\uparrow$6.66} 
      & \cellcolor{lightgrayv}\textbf{83.33}{\color{myred} \scriptsize $\uparrow$6.67} 
      & \cellcolor{lightgrayv}53.33{\color{myred} \scriptsize $\uparrow$10.83} 
      & \cellcolor{lightgrayv}76.66{\color{myred} \scriptsize $\uparrow$10.00} 
      & \cellcolor{lightgrayv}90.60{\color{myred} \scriptsize $\uparrow$5.00} 
      & \cellcolor{lightgrayv}57.18{\color{myred} \scriptsize $\uparrow$7.92} 
      & \cellcolor{lightgrayv}66.48 \\
    & \cellcolor{lightgrayv}+ \babyb {\scriptsize\ (ours)}
      & \cellcolor{lightgrayv}\textbf{68.33}{\color{myred} \scriptsize $\uparrow$10.83} 
      & \cellcolor{lightgrayv}\textbf{83.33}{\color{myred} \scriptsize $\uparrow$6.67} 
      & \cellcolor{lightgrayv}\textbf{55.00}{\color{myred} \scriptsize $\uparrow$12.50} 
      & \cellcolor{lightgrayv}\textbf{80.00}{\color{myred} \scriptsize $\uparrow$13.34} 
      & \cellcolor{lightgrayv}\textbf{92.20}{\color{myred} \scriptsize $\uparrow$6.60} 
      & \cellcolor{lightgrayv}\textbf{57.64}{\color{myred} \scriptsize $\uparrow$8.38} 
      & \cellcolor{lightgrayv}\textbf{68.16} \\
    \hline
    % 7B-Instruct
    \multirow{4}{*}{\rotatebox{90}{\fontsize{8pt}{10pt} \textbf{7B-Instruct}\ }}
    & + GRACE {\scriptsize \citep{zhang2025the}}
      & 15.00 & 36.66 & 15.00 & 26.66 & 73.80 & 34.85 & 35.08 \\
    & + Local LP {\scriptsize \citep{just2025distilling}} 
      & 17.50 & 36.66 & 16.66 & 26.66 & 74.60 & 35.58 & 35.81 \\
    & \cellcolor{lightgrayv}+ \babya {\scriptsize\ (ours)}
      & \cellcolor{lightgrayv}25.00{\color{myred} \scriptsize $\uparrow$7.50} 
      & \cellcolor{lightgrayv}43.33{\color{myred} \scriptsize $\uparrow$6.67} 
      & \cellcolor{lightgrayv}22.50{\color{myred} \scriptsize $\uparrow$5.84} 
      & \cellcolor{lightgrayv}40.00{\color{myred} \scriptsize $\uparrow$13.34} 
      & \cellcolor{lightgrayv}82.00{\color{myred} \scriptsize $\uparrow$7.40} 
      & \cellcolor{lightgrayv}41.61{\color{myred} \scriptsize $\uparrow$6.03} 
      & \cellcolor{lightgrayv}42.35 \\
    & \cellcolor{lightgrayv}+ \babyb {\scriptsize\ (ours)}
      & \cellcolor{lightgrayv}\textbf{30.00}{\color{myred} \scriptsize $\uparrow$12.50} 
      & \cellcolor{lightgrayv}\textbf{50.00}{\color{myred} \scriptsize $\uparrow$13.34} 
      & \cellcolor{lightgrayv}\textbf{24.16}{\color{myred} \scriptsize $\uparrow$7.50} 
      & \cellcolor{lightgrayv}\textbf{43.33}{\color{myred} \scriptsize $\uparrow$16.67}
      & \cellcolor{lightgrayv}\textbf{82.40}{\color{myred} \scriptsize $\uparrow$7.80} 
      & \cellcolor{lightgrayv}\textbf{46.32}{\color{myred} \scriptsize $\uparrow$10.74} 
      & \cellcolor{lightgrayv}\textbf{46.07} \\
    \bottomrule
  \end{tabular} }
\end{table*}

\subsection{Performance on Alleviating Confounding}

To examine whether our proposed methods effectively address the step length confounding problem, we present in Fig.~\ref{step_length_ours} the distributions of step lengths for data samples \textit{selected} and \textit{unselected} by our two variants \babya and \babyb. 
In contrast to the significant differences in step length distributions exhibited by prior approaches in Fig.~\ref{mix_teacher_bias}, our approaches yield markedly smaller step length disparities. This demonstrates that our methods successfully mitigate step length confounding. Moreover, because both variants operate by intervening directly on the probability assigned to the first token, this result further implies that the step length confounding issue is intimately linked to the model’s first-token probabilities.

\subsection{Comparing Min and Max Probabilities}

We also compare the performance of the \babyb variant when selecting samples with the highest ($\max$ \textsc{Casl}) versus the lowest ($\min$ \textsc{Casl}) probability values of metric $s^{\mathrm{casl}}$. All models are trained on the \textit{LIMO‑v2} dataset, and Table~\ref{minmax} reports their performance on four evaluation benchmarks after training the four target LLMs.
The experimental results consistently show that selecting samples with the highest probabilities outperforms selecting the lowest ones, a finding aligned with previous naturalness‑based selection methods. In general, high-probability samples correspond to data that better align with the target LLM’s current capabilities, enabling more effective and stable learning of the SFT data distribution, thereby leading to superior performance. Conversely, lower-probability samples reflect that the model is less familiar with the data, which can introduce noisy training gradients and ultimately degrade model performance.

\begin{table}[t]
\centering
\renewcommand\arraystretch{1.05}
  \caption{Comparison of the experimental results for \babyb that selects the highest and lowest $s^{\mathrm{casl}}$.}
  \label{minmax}
  \small
  \setlength{\tabcolsep}{5pt}{
  \begin{tabular}{m{2.0cm}<{\centering}m{1.0cm}<{\centering}m{1.0cm}<{\centering}m{0.95cm}<{\centering}m{0.95cm}<{\centering}}
    \toprule
    Method & AIME24 & AIME25 & MATH & OlymB. \\
    \hline
    \multicolumn{5}{c}{\fontsize{8pt}{11pt}\selectfont  \textbf{Qwen3-4B-Base}} \\
    \rowcolor{lightgrayv} $\max$ \textsc{Casl} & 31.66 & 30.83 & 80.00 & 42.81 \\
    $\min$ \textsc{Casl} & 29.16 & 28.33 & 77.40 & 39.70 \\
    \hline
    \multicolumn{5}{c}{\fontsize{8pt}{11pt}\selectfont  \textbf{Qwen3-8B-Base}} \\
    \rowcolor{lightgrayv} $\max$ \textsc{Casl} & 45.00 & 37.50 & 85.40 & 49.03 \\
    $\min$ \textsc{Casl} & 41.66 & 36.66 & 79.60 & 42.94 \\
    \hline
    \multicolumn{5}{c}{\fontsize{8pt}{11pt}\selectfont  \textbf{Qwen3-4B-Instruct}} \\
    \rowcolor{lightgrayv} $\max$ \textsc{Casl} & 71.66 & 58.33 & 93.20 & 60.44 \\
    $\min$ \textsc{Casl} & 65.83 & 55.83 & 86.80 & 55.14 \\
    \hline
    \multicolumn{5}{c}{\fontsize{8pt}{11pt}\selectfont  \textbf{Qwen2.5-7B-Instruct}} \\
    \rowcolor{lightgrayv} $\max$ \textsc{Casl} & 28.33 & 24.16 & 81.60 & 45.18 \\
    $\min$ \textsc{Casl} & 25.00 & 22.50 & 79.60 & 43.08 \\
    \bottomrule
  \end{tabular} }
\end{table}

\subsection{Linear Regression Results} \label{sec:regression}

Our method \babyb fits a linear regression model as defined in Eq.~\eqref{eq5}, and uses this model to remove the influence of the first‑token ratio on the average log probability. Accordingly, Table~\ref{regression} presents the fitted results of the linear regression for data originating from different source LLMs.
First, $\gamma$ is the most important coefficient, as it directly determines the extent to which the first‑token ratio affects the average probability. Overall, the largest $\gamma$ value among the models is -1.284, meaning that for every 0.05 difference in the first‑token ratio between samples, the impact on the overall probability is equivalent to reducing each token’s probability by $1 -  e^{-1.284 \times 0.05} = 6.22\%$. For the regression fitted on all SFT data, $\gamma$ is -0.680, corresponding to a $1 -  e^{-0.680 \times 0.05} = 3.34\%$ reduction in per‑token probability.
Comparing different source LLMs, \textit{gpt-oss-120b} exhibits the largest $\gamma$ value, indicating that the data from it suffers from a more pronounced confounding problem.

In contrast, when comparing $\beta_1$ (the first‑token probability) and $\beta_2$ (the non‑first‑token probability), we observe $\beta_1 \ll \beta_2$, which further suggests that the ratio of the first‑token probability in the global average probability should be reduced.
Lastly, $\epsilon$ consistently remains at a low level, implying that the regression model has only minor fitting bias, and thus the debiasing results are accurate.

\begin{table}[t]
\centering
\renewcommand\arraystretch{1.05}
  \caption{Linear regression parameters of Eq.~\eqref{eq5} fitted on data generated by different source LLMs on \textit{LIMO‑v2}.}
  \label{regression}
  \small
  \setlength{\tabcolsep}{5pt}{
  \begin{tabular}{m{2.0cm}<{\centering}m{1.0cm}<{\centering}m{1.0cm}<{\centering}m{0.95cm}<{\centering}m{0.95cm}<{\centering}}
    \toprule
    Source & $\beta_1$ & $\beta_2$ & $\gamma$ & $\epsilon$ \\
    \hline
    \textit{QwQ-32B} & 0.077 & 0.919 & \cellcolor[HTML]{F4F3F8} -0.284 & -0.001 \\
    \textit{Qwen3-32B} & 0.068 & 0.929 & \cellcolor[HTML]{F0EFF4} -0.529 & 0.007 \\
    \textit{DS-Qwen-32B} & 0.066 & 0.943 & \cellcolor[HTML]{FFFFFF} -0.226 & 0.001 \\
    \textit{gpt-oss-120b} & 0.068 & 0.938 & \cellcolor[HTML]{ECEBF0} \textbf{-1.284} & 0.028 \\
    \hline
    \rowcolor{lightgrayv} Overall & 0.066 & 0.944 & -0.680 & -0.054 \\
    \bottomrule
  \end{tabular} }
\end{table}

\subsection{Computation Budget}

Compared to GRACE, our \babya variant adopts a more streamlined approach: it simply drops the first token when computing the average token probability, thereby avoiding any additional computational overhead. In contrast, our \babyb variant introduces a modest amount of extra computation by fitting a lightweight linear regression model. However, since this regression model involves only a small number of parameters, the fitting procedure is highly efficient and typically completes in just a few seconds, imposing negligible cost on the overall pipeline.

%% file: S_RelatedWorks.tex
\section{Related Works}

Recently, instead of direct prompt LLMs to generate CoT responses \citep{wei2022chain,yuan2024instance,bi2025cot}, leveraging SFT to elicit long CoT reasoning in LLMs has emerged as a standard training paradigm, outperforming large-scale reinforcement learning even when applied to smaller models \citep{guo2025deepseek,yang2025qwen3,tian2025reinforcement,kou2026positive}.
Generally, the existing methods typically scale up the SFT data generated by a strong LLM by constructing a large and diverse set of questions \citep{zhao20251,guha2025openthoughts,yuan2025naturalreasoning}, and generating diverse solutions for each question through temperature sampling \citep{chen2023mcc,lei2025learning,chen2025acereason,yan2026distribution}.
Building on these large-scale datasets, some studies have also sought to filter out noisy data by applying various heuristic rules, \eg question difficulty \citep{muennighoff2025s1,guha2025openthoughts,li2025naturalthoughts}, solution correctness \citep{chen2025skip,luo2025deconstructing}, diversity of solutions \citep{jung2025prismatic,li2025exploring}, and LLM-as-a-Judge \citep{wu2025beyond,lei2025learning}. Remarkably, even reducing the dataset to only 1k questions can still elicit the long CoT reasoning capability \citep{muennighoff2025s1,ye2025limo}.

Beyond heuristic data‑selection strategies, several works advocate naturalness‑based approaches \citep{zhang2025the,just2025distilling,liu2026where}, wherein data are selected based on the model’s confidence scores, allowing the selection of samples to which the model is better adapted.
Although naturalness‑based methods can indeed assess a model’s adaptability to data via confidence scores \citep{kang2025scalable,fu2025deep}, recent studies have shown that reasoning data often contain a small number of high‑entropy / low-probability first tokens \citep{yang2025do,wang2025beyond}. Our experiments further confirm that these low-probability first tokens can substantially exacerbate the step‑length confounding problem in naturalness‑based approaches.

%% file: S_Conclusion.tex
\section{Conclusion}

In this work, we investigate the limitations of naturalness‑based data selection for long CoT reasoning datasets. Our analysis reveals a systematic bias, termed step length confounding, whereby the selection pipeline significantly prefers samples with longer reasoning steps instead of those with higher reasoning quality. We trace this phenomenon to the disproportionate influence of low‑probability first tokens in reasoning steps, which is diluted in longer sequences, thus inflating their average log probabilities.
To mitigate this problem, we propose \babya and \babyb, two variants that drop or causally debias the first‑token probability when computing selection scores. Extensive experiments on two reasoning SFT datasets, across four LLMs and five evaluation benchmarks, demonstrate that our approaches consistently outperform existing naturalness-based selection methods and effectively alleviate the step length confounding problem.

\section*{Limitations}

This paper systematically investigates an important property in SFT data for LLM reasoning: the relationship between response and step length, and naturalness-based data selection. From a methodological perspective, one major limitation lies in our identification of the influence of a critical first token on step length confounding; whether there are deeper confounding factors remains an open question. 
In addition, recent studies have focused on on-policy data generation and selection \citep{yang2025qwen3,lu2025on}, where the student model produces its own samples for training. Whether data selection in such approaches still exhibits a strong correlation with response length is an issue worthy of further exploration.

\section*{Acknowledgement}

This work was supported in part by the National Natural Science Foundation of China (No.62276113).

%% file: S_Appendix.tex
\appendix

\section{Bias Towards Response Length} \label{sec:app.total}

In our preliminary experiments, we also observe that, under the same setup as in Fig.~\ref{mix_teacher_bias}, the average total response length (\ie the total token number in the full response) of data \textit{selected} by naturalness-based methods is approximately 9.8K, compared to about 15.4K for \textit{unselected} data, revealing a significant discrepancy.
Therefore, in the following sections, we aim to address the following question through experiments: \textit{are samples with shorter overall response lengths more likely to be selected by naturalness-based data selection methods}?

\subsection{Longer Response, Higher Log Probability}  \label{sec:app.total.longer}
\begin{figure}[t]
  \centering
  \includegraphics[width=1.0\columnwidth]{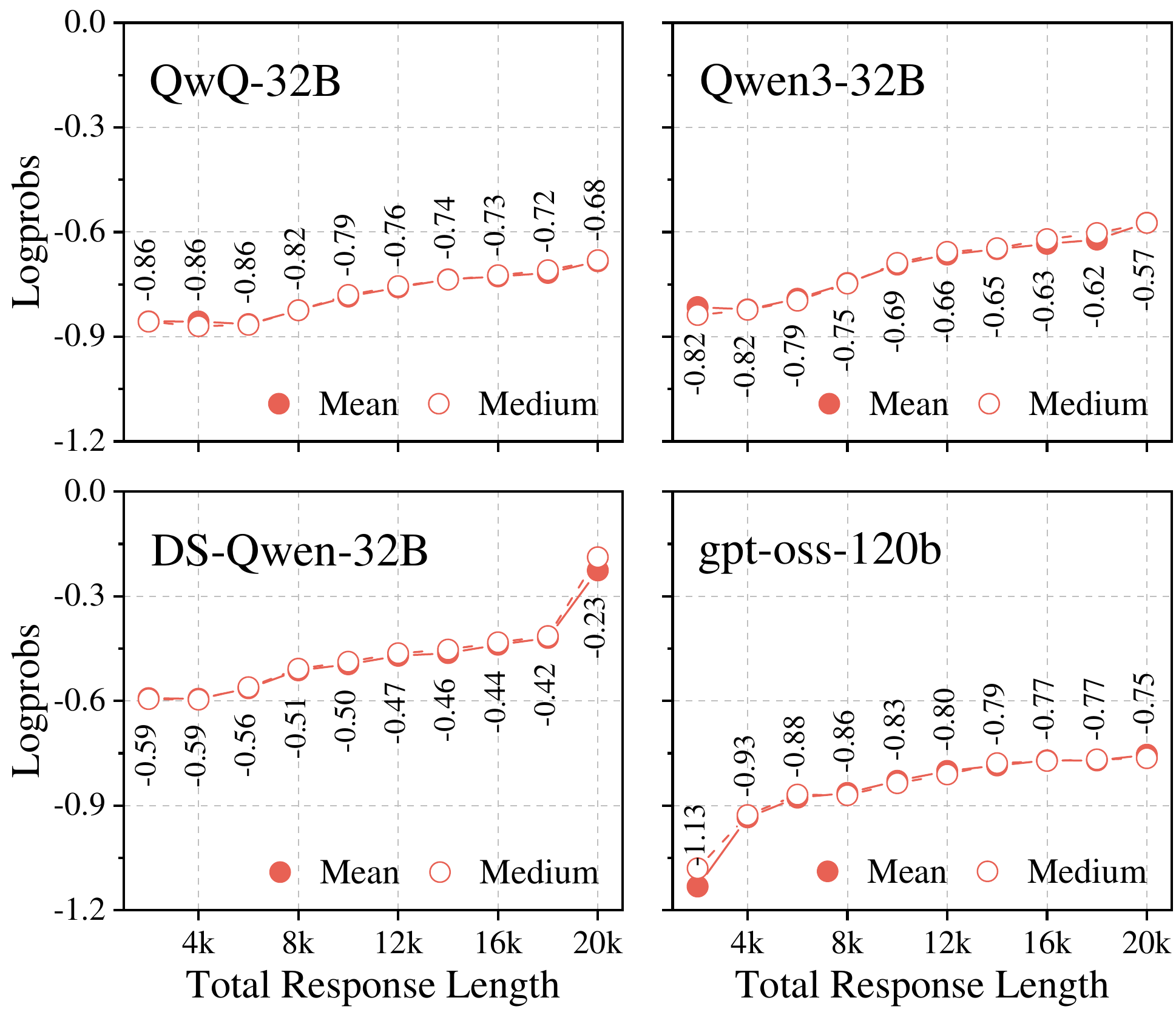}
  \caption{Relationship between response-level log probability and total response length.}
  \label{total_length_logits}
  \vspace{-2pt}
\end{figure}

First, we maintain the same experimental setup as in Sec.~\ref{sec.2.2}, using \textit{Qwen3-4B-Base} as the target model to compute the log probabilities for data generated by four different LLMs. Fig.~\ref{total_length_logits} illustrates the trend of globally averaged log probabilities (\ie $s_i^{\mathrm{logp}}$) \wrt overall response length. The results show that, as the response length increases, the log probabilities actually rise, \textit{contrary to the earlier conclusion, where shorter responses were more likely to be selected}. 
Comparing different models, we find that the average log probabilities of \textit{DS-Qwen-32B}, which exhibits longer step lengths, are consistently and substantially higher than those of other models; even the highest log probability among other models is lower than the lowest value from \textit{DS-Qwen-32B}. In summary, although log probabilities should in principle increase with response length, the selection bias caused by step length confounding has a much stronger influence than the effect of overall response length, leading to the seemingly opposite conclusion above.

\subsection{Why Response Length Bias?}

\begin{figure}[t]
  \centering
  \includegraphics[width=0.87\columnwidth]{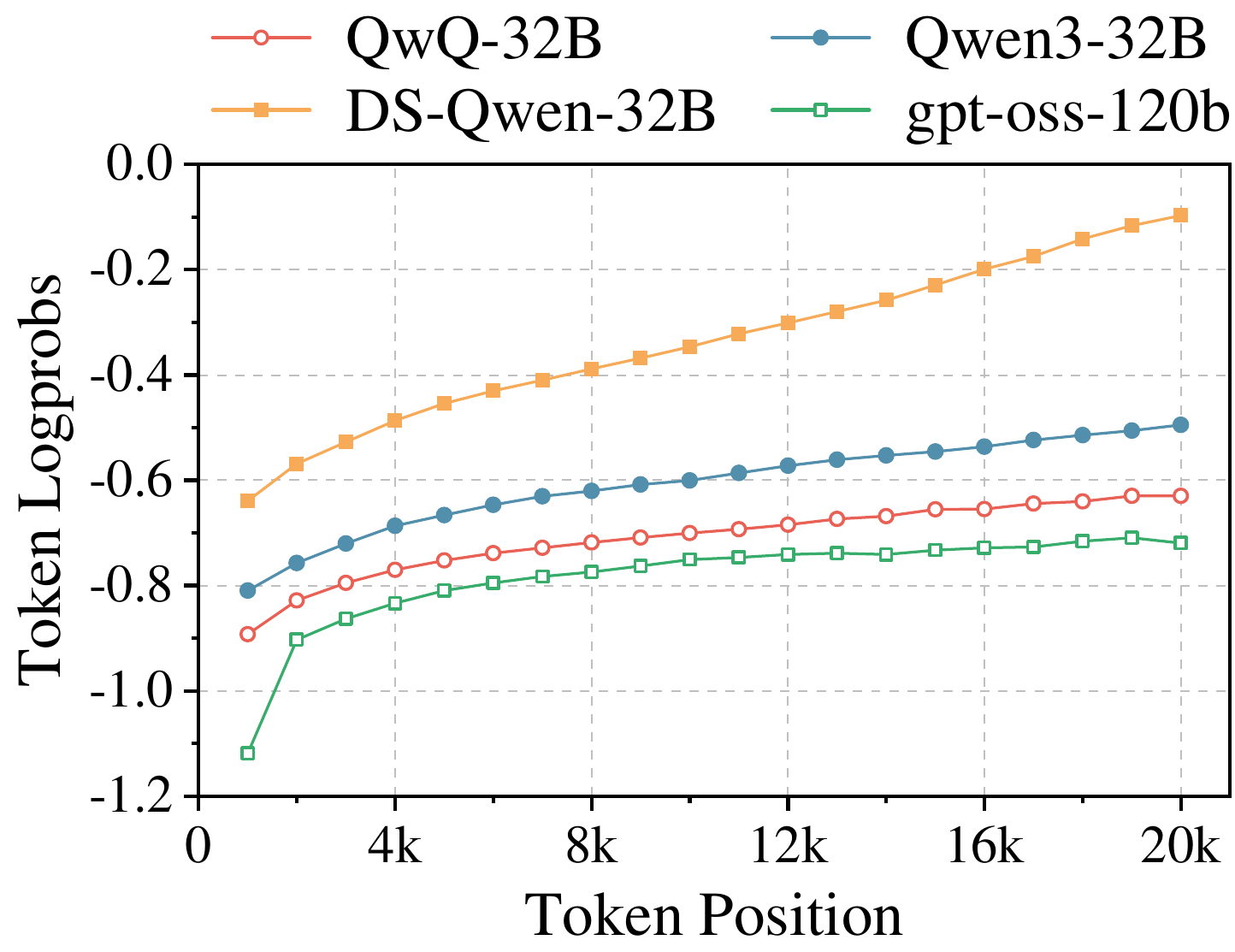}
  \caption{Average log probability of tokens at different positions for four source LLMs.}
  \label{logprob_avg}
\end{figure}

This section aims to further investigate why longer responses tend to have higher log probabilities. Fig.~\ref{logprob_avg} presents the log probabilities of tokens at different positions within the responses. Specifically, we first determine the 95th percentile of the maximum response length, and then divide the range from 0 to this value into 20 bins. Tokens are assigned to these bins according to their position within the response, and the average log probability is computed for the tokens in each bin. As shown in Fig.~\ref{logprob_avg}, there is a clear trend: \textit{tokens located toward the end of a response have higher log probabilities than those at the beginning}, following a monotonically increasing pattern. This is because, as the response length grows, the target LLM often becomes more confident in generating the continuation, indicating better adaptation to the data. Consequently, longer responses exhibit higher log probabilities for their tail-end tokens, which in turn results in a higher overall log probability.

\subsection{Step Length Significantly Matters Response Length for Data Selection}
In Sec.~\ref{sec:app.total.longer}, we present the experimental finding that the bias induced by total response length is negligible compared to the step length confounding issue. To further validate this conclusion, we employ the causal regression approach proposed in Sec.~\ref{sec.3.2}, rewriting Eq.~\eqref{eq5} as follows:
\begin{equation}
    s_i^{\mathrm{logp}} = \beta_1 s_i^{\mathrm{first}} + \beta_2 s_i^{\mathrm{drop}} + \gamma \mathcal{Z}_i + \gamma_2 |\mathbf{o}_i| + \epsilon. \nonumber
\end{equation}
Using the same experimental settings, we refit the model, and the final parameter estimation results are reported in Table~\ref{app.regression}. Compared with the confounder $\gamma \mathcal{Z}_i$ introduced by step length, the confounder $\gamma_2 |\mathbf{o}_i|$ induced by total length is smaller by approximately two orders of magnitude.\footnote{$\mathcal{Z}_i$ is typically on the order of $10^{-1}$, whereas $|\mathbf{o}_i|$ is typically on the order of $10^5$.}

In Table~\ref{remove_total_length}, we further compare the performance metrics of models trained on data selected \textit{with} and \textit{without} the $\gamma_2 |\mathbf{o}_i|$ term in the criterion, \ie $s_i^{\mathrm{casl}} = s_i^{\mathrm{logp}} - \gamma \mathcal{Z}_i - \gamma_2 |\mathbf{o}_i|$. The results show that removing the $\gamma_2 |\mathbf{o}_i|$ term causes little change in model performance, once again confirming that the influence of total length is generally small and can even be negligible. In fact, prior studies have provided evidence that longer reasoning SFT data \citep{chen2025acereason,guha2025openthoughts} or in-context CoT prompts \citep{jin2024the} can be more effective for improving model performance. This suggests that retaining the bias associated with total response length might even be beneficial. However, the step length confounding phenomenon leads to the opposite outcome, preferring shorter responses, which contradicts these findings and further underscores the importance of mitigating this bias.

\begin{table}[t]
\centering
\renewcommand\arraystretch{1.0}
  \caption{Linear regression parameters including overall response length $\gamma_2 |\mathbf{o}_i|$.}
  \label{app.regression}
  \small
  \setlength{\tabcolsep}{5pt}{
  \begin{tabular}{m{2.0cm}<{\centering}m{0.65cm}<{\centering}m{0.65cm}<{\centering}m{0.85cm}<{\centering}m{0.8cm}<{\centering}m{0.7cm}<{\centering}}
    \toprule
    Source & $\beta_1$ & $\beta_2$ & $\gamma$ & $\gamma_2$ & $\epsilon$ \\
    \hline
    \textit{QwQ-32B} & .028 & .972 & \cellcolor{lightgrayv} -0.062 & \cellcolor{lightgrayv} 1E-8 & .004 \\
    \textit{Qwen3-32B} & .018 & .979 & \cellcolor{lightgrayv} -0.043 & \cellcolor{lightgrayv} 3E-9 & .007 \\
    \textit{DS-Qwen-32B} & .014 & .983 & \cellcolor{lightgrayv} -0.158 & \cellcolor{lightgrayv} 4E-8 & -.008 \\
    \textit{GPT-OSS-120B} & .018 & .987 & \cellcolor{lightgrayv} -1.166 & \cellcolor{lightgrayv} 5E-9 & .025 \\
    \bottomrule
  \end{tabular} }
\end{table}

\begin{table}[t]
\centering
\renewcommand\arraystretch{1.0}
  \caption{Comparison of experimental results with and without removing overall response length bias $\gamma_2 |\mathbf{o}_i|$.}
  \label{remove_total_length}
  \small
  \setlength{\tabcolsep}{5pt}{
  \begin{tabular}{m{2.0cm}<{\centering}m{1.0cm}<{\centering}m{1.0cm}<{\centering}m{0.95cm}<{\centering}m{0.95cm}<{\centering}}
    \toprule
    Teacher & AIME24 & AIME25 & MATH & OlymB. \\
    \hline
    \multicolumn{5}{c}{\textbf{Qwen3-4B-Base}} \\
    \rowcolor{lightgrayv} \babyb & 31.66 & 30.83 & 80.00 & 42.81 \\
     + $\gamma_2 |\mathbf{o}_i|$ & 30.83 & 30.83 & 78.60 & 42.20 \\
    \hline
    \multicolumn{5}{c}{\textbf{Qwen3-8B-Base}} \\
    \rowcolor{lightgrayv} \babyb & 45.00 & 37.50 & 85.40 & 49.03 \\
     + $\gamma_2 |\mathbf{o}_i|$ & 43.33 & 35.83 & 83.80 & 48.52 \\
    \bottomrule
  \end{tabular} }
\end{table}

\section{More Experimental Settings} \label{sec:appendix.settings}

In this section, we provide a detailed description of our experimental settings, including LLM model cards, data processing pipelines, and SFT details.

\subsection{LLM Model Cards} \label{sec:appendix.settings.model}

In our experiments, we employ two categories of LLMs: those used to generate SFT data, which we refer to as \textit{source LLMs}, and those trained on the generated SFT data, which we refer to as \textit{target LLMs}. Their details are described as follows.

\vspace{2pt} \noindent
\textbf{Source LLMs.}
We use four different families of LLMs, each producing five distinct responses for every question.
\begin{itemize}
    \item \textit{QwQ-32B}\footnote{\fontsize{8.0}{10}\selectfont \url{https://huggingface.co/Qwen/QwQ-32B}} \citep{qwen2025qwq32b} is a specialized reasoning model trained with reinforcement learning on top of \textit{Qwen2.5‑32B}.
    \item \textit{Qwen3-32B}\footnote{\fontsize{8.0}{10}\selectfont \url{https://huggingface.co/Qwen/Qwen3-32B}} \citep{yang2025qwen3} has undergone large-scale long-CoT cold-start training and reasoning-focused reinforcement learning. In its technical report, this model is used to distill smaller-scale models, \eg the \textit{Qwen3-4B} and \textit{Qwen3-8B} variants, which align with our experimental setup.
    \item \textit{DeepSeek-R1-Distill-Qwen-32B}\footnote{\fontsize{8.0}{10}\selectfont \url{https://huggingface.co/deepseek-ai/DeepSeek-R1-Distill-Qwen-32B}} \citep{guo2025deepseek} is one of the first to employ reinforcement learning to enhance long CoT reasoning in LLMs, providing evidence that models obtained through distillation can still exhibit robust reasoning abilities.
    \item \textit{gpt-oss-120b}\footnote{\fontsize{8.0}{10}\selectfont \url{https://huggingface.co/openai/gpt-oss-120b}} \citep{agarwal2025gpt} improves inference speed by combining compact attention layers with linear attention layers, while activating only 5B parameters. The model is also trained using the conventional paradigm of SFT followed by reinforcement learning.
\end{itemize}

\vspace{2pt} \noindent
\textbf{Target LLMs.}
Using the generated SFT data, we train four LLMs of varying sizes and types.

\begin{itemize}
    \item \textit{Qwen3-4B-Base}\footnote{\fontsize{8.0}{10}\selectfont \url{https://huggingface.co/Qwen/Qwen3-4B-Base}} and \textit{Qwen3-8B-Base}\footnote{\fontsize{8.0}{10}\selectfont \url{https://huggingface.co/Qwen/Qwen3-8B-Base}} are two different-sized models that have undergone large-scale pre-training only.
    \item \textit{Qwen3-4B-Instruct}\footnote{\fontsize{8.0}{10}\selectfont \url{https://huggingface.co/Qwen/Qwen3-4B-Instruct-2507}} and \textit{Qwen2.5-7B-Instruct}\footnote{\fontsize{8.0}{10}\selectfont \url{https://huggingface.co/Qwen/Qwen2.5-7B-Instruct}} build upon the two base versions described above, undergoing thorough instruction fine-tuning. For the 4B model, we adopt its 2507 variant, updating the non-thinking mode from \textit{Qwen3-4B}’s mixed-reasoning framework. Furthermore, as the \textit{Qwen3} series lacks an instruct model of 8B parameters, we use the 7B instruct model from the \textit{Qwen2.5} series as a substitute.
\end{itemize}

\subsection{Data Sampling and Filtering} \label{sec:appendix.settings.data}

Our experiments are conducted using datasets from two different sources.

\begin{itemize}
    \item \textit{LIMO-v2}\footnote{\fontsize{8.0}{10}\selectfont \url{https://huggingface.co/datasets/GAIR/LIMO-v2}} \citep{ye2025limo} undergoes rigorous quality filtering, resulting in a final selection of 800 high-quality mathematics problems. For this dataset, we generate five diverse correct responses for each problem using every source LLM.
    \item \textit{AceReason-1.1-SFT}\footnote{\fontsize{8.0}{10}\selectfont \url{https://huggingface.co/datasets/nvidia/AceReason-1.1-SFT}} \citep{chen2025acereason} aggregates large-scale, high-quality SFT data from multiple sources. From this dataset, we randomly sample 10k mathematics problems, and for each problem, we obtain one correct response generated by each of the four aforementioned source LLMs.
\end{itemize}

For these two datasets, we adopt the following data sampling and quality filtering pipeline.

\vspace{2pt} \noindent
\textbf{Data sampling.}
During data sampling, we employ top‑$p$ sampling with $p$ set to 0.95. For gpt-oss-120b, following the official recommendations, we set the sampling temperature to 1.0 and the reasoning effort to high. For the other source LLMs, the sampling temperature is set to 0.6. All data sampling is conducted via SGLang for offline LLM deployment and calling, with the maximum generation length consistently fixed at 64K tokens.

\vspace{2pt} \noindent
\textbf{Data filtering.}
We perform dynamic filtering during the sampling process. Specifically, for each problem, we sample one response at a time and verify it using the Math‑Verify toolkit.\footnote{\fontsize{8.0}{10}\selectfont \url{https://github.com/huggingface/Math-Verify}} Sampling continues until the required number of correct responses is obtained (5 responses for \textit{LIMO‑v2} and 1 response for \textit{AceReason-1.1-SFT}), or until the number of attempts exceeds 15. 
In practice, some problems are difficult to collect five correct responses for, so we repeat the above procedure five times to ensure that every problem has sufficient correct responses. 
For problems that are excessively difficult and fail to meet the required number of correct responses, we adopt the following remedies. In \textit{LIMO‑v2}, we manually sample several responses that are close to the correct answer and supplement them until five correct responses are obtained. In \textit{AceReason-1.1-SFT}, we sample additional problems from the dataset and continue generation until we reach a total of 10k problems, each paired with its corresponding correct response.
As a result, each problem in \textit{LIMO‑v2} may contain up to 75 incorrect responses and at least 25 correct ones. All generated response data are publicly available via the link provided in this paper.

\begin{figure}[t]
  \centering
  \includegraphics[width=\columnwidth]{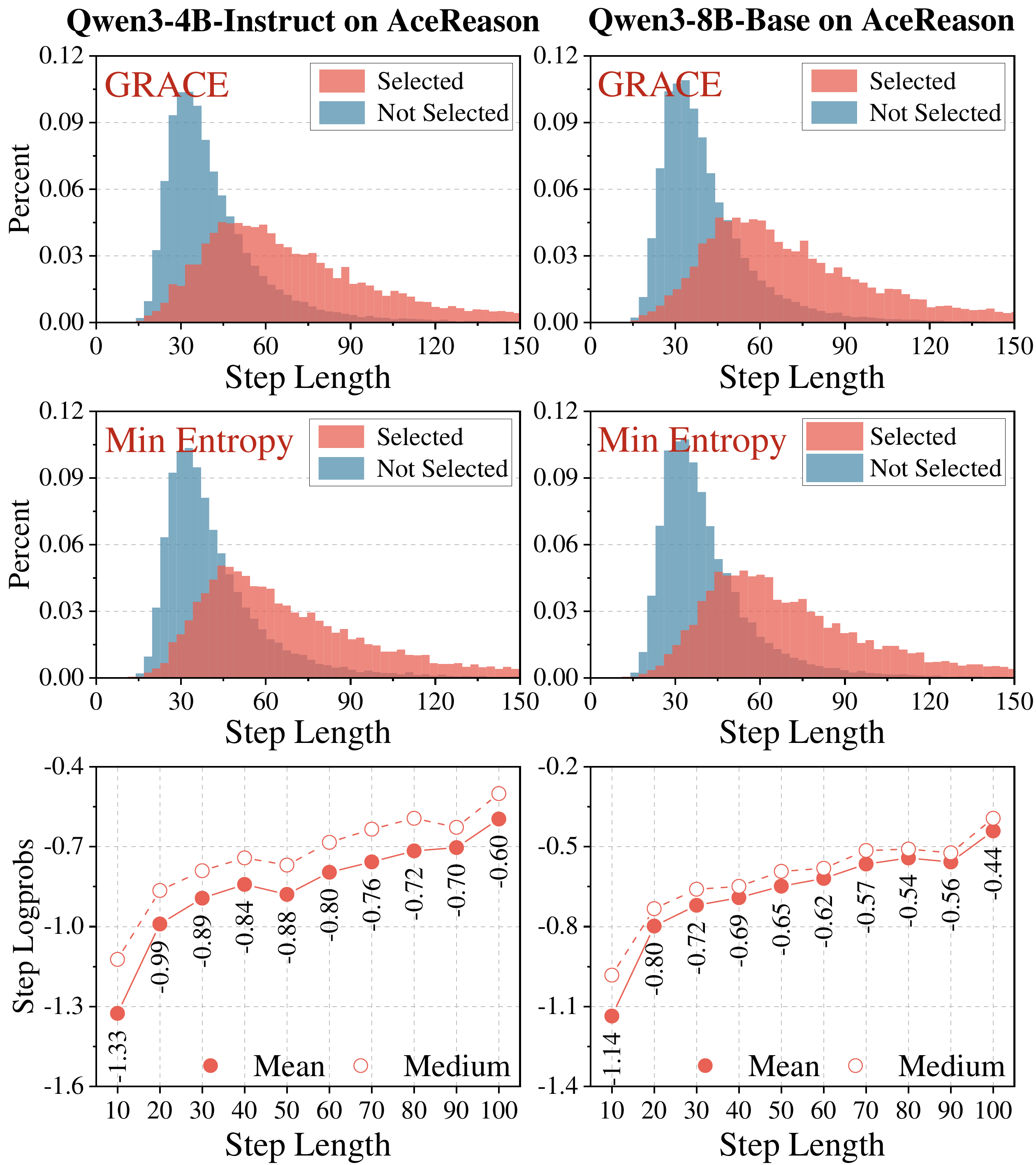}
  \caption{Step length distributions for \textit{selected} and \textit{unselected} data and relationship between step-level log probability and step length on \textit{AceReason‑1.1‑SFT}.}
  \label{step_length_logits_acereason}
\end{figure}

\vspace{2pt} \noindent
\textbf{Sampling log probabilities of target LLMs.}
We employ SGLang to sample log probabilities from the target LLM for the data.
In GRACE \citep{zhang2025the}, the output log probabilities are averaged directly.
In Local LP \citep{just2025distilling}, following the best practices outlined in the original paper, each problem and step is paired with its preceding $k = 4$ steps, the average log probability is computed for each step, and the step‑level averages are then averaged.
For Min Entropy, retrieving vocabulary‑level probability distributions for every token is computationally and storage‑intensive. Given their typical long‑tailed nature, we instead retain only the top 20 most probable tokens per token, compute the entropy for each, and average these values across all tokens in a response.

\subsection{Fine-tuning Details}
Using the reasoning SFT data, we fine‑tune the target LLM with full‑parameter training via LlamaFactory. We set the training batch size to 32 and enable LlamaFactory’s built‑in \textit{packing} option to concatenate shorter samples, with the maximum sequence length fixed at 32K. Optimization is carried out using the Adam optimizer with a learning rate of $5 \times 10^{-5}$ for a total of 6 epochs. The learning rate scheduler is configured as \textit{cosine\_with\_min\_lr}, with a minimum learning rate of $1 \times 10^{-5}$.

\section{More Analysis Results} \label{sec:appendix.analysis}

In this section, we provide additional analytical results regarding step length confounding.

\subsection{Analysis on More Datasets and Models} 

In Sec.~\ref{sec.preliminary}, we analyze the \textit{LIMO‑v2} dataset using \textit{Qwen3‑4B‑Base} as the target LLM, examining results across four different source LLMs. In this section, we extend our analysis to the \textit{AceReason‑1.1‑SFT} dataset and combine data from all source LLMs, evaluating their performance on two target models: \textit{Qwen3‑4B‑Instruct} and \textit{Qwen3‑8B‑Base}. The analysis results are shown in Fig.~\ref{step_length_logits_acereason}.
The experimental results show that, for both target LLMs, the \textit{AceReason‑1.1‑SFT} dataset exhibits a pronounced difference in step‑length distribution between the \textit{selected} and \textit{unselected} samples. Moreover, the monotonic increasing relationship between each step’s log probability and its length remains significant, with the probabilities from \textit{Qwen3‑8B‑Base} consistently exceeding those of \textit{Qwen3‑4B‑Instruct}.

\begin{figure}[t]
  \centering
  \includegraphics[width=1.0\columnwidth]{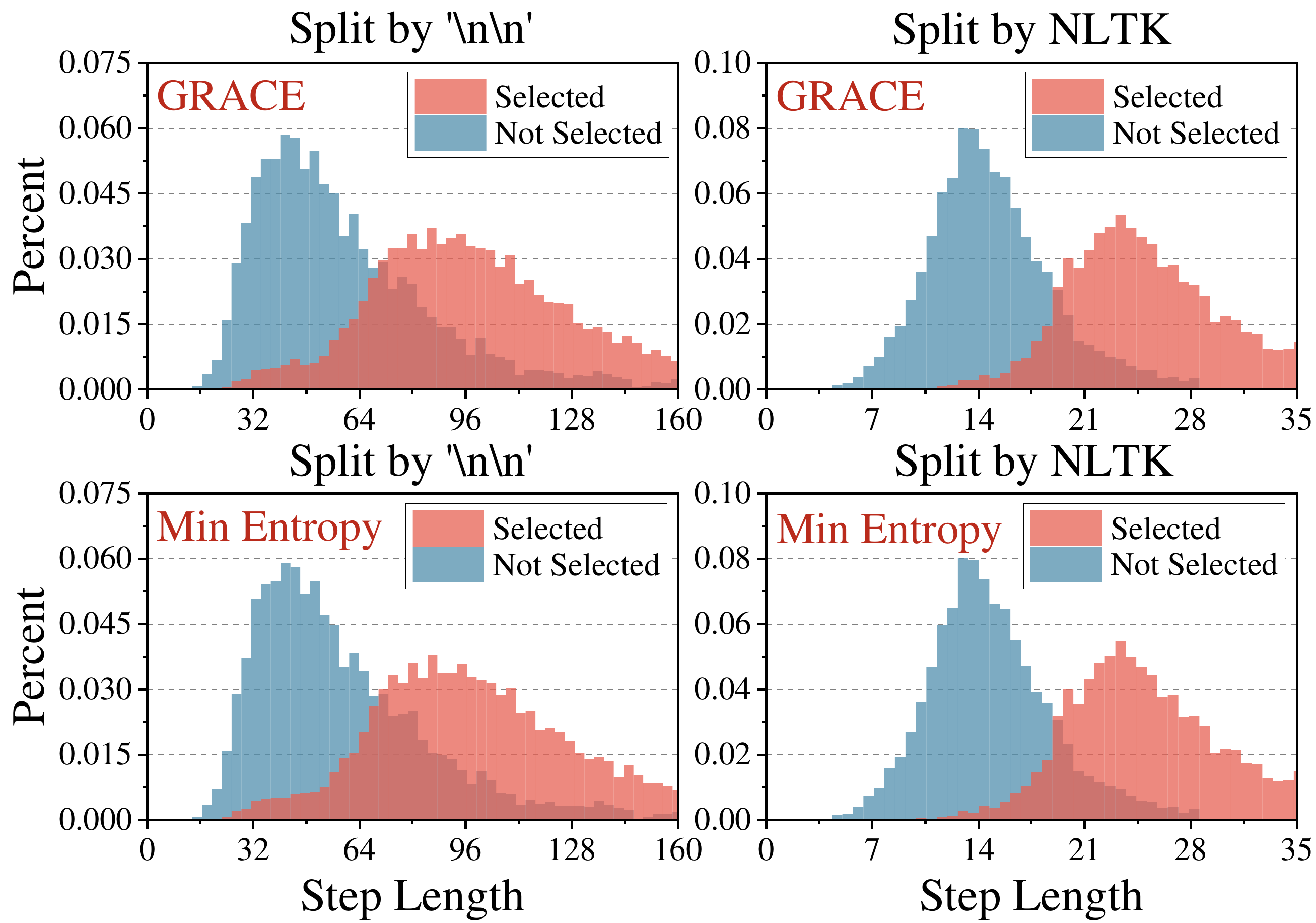}
  \caption{Data selection bias and step length distributions under different splitting methods.}
  \label{mix_teacher_bias_different_split}
\end{figure}

\subsection{Different Step Splitting Methods}

The step-length distribution differences shown in Fig.~\ref{mix_teacher_bias} are obtained by splitting the responses into steps using periods and spaces. In the community, aside from this splitting approach, some studies adopt \texttt{\textbackslash n\textbackslash n} or external tools such as NLTK for step splitting. Therefore, we also investigate the step length‑distribution differences under these two additional step splitting methods. The analysis results are presented in Fig.~\ref{mix_teacher_bias_different_split}.

Our experimental results consistently show that, regardless of the step splitting method used, the \textit{selected} versus \textit{unselected} data still display a clear difference in step‑length distribution. This indicates that the naturalness‑based data selection approach continues to suffer from a step length confounding issue. Moreover, splitting sentences using periods produces the most distinct distribution differences compared to other splitting methods. This leads to another key observation: low-probability tokens are most prominent at sentence beginnings when period‑based splitting is applied, as opposed to the first tokens in steps segmented by other methods.

\subsection{Casual Regression Parameters}

We apply the causal regression method introduced in Sec.~\ref{sec.3.2} to the \textit{AceReason‑1.1‑SFT} dataset, re‑fitting the two target LLMs analyzed previously, and present the regression parameters in Table~\ref{regression_acereason}. The results show that the $\gamma$ values remain high, indicating that the step length confounding issue persists. Furthermore, the result $\beta_1 \ll \beta_2$ is fully consistent with the conclusions presented in Sec.~\ref{sec:regression}.

\begin{table}[t]
\centering
\renewcommand\arraystretch{1.0}
  \caption{Linear regression parameters of all SFT data on \textit{AceReason‑1.1‑SFT} for the two target models.}
  \label{regression_acereason}
  \small
  \setlength{\tabcolsep}{5pt}{
  \begin{tabular}{m{2.45cm}<{\centering}m{0.83cm}<{\centering}m{0.83cm}<{\centering}m{0.83cm}<{\centering}m{0.83cm}<{\centering}}
    \toprule
    Target LLM & $\beta_1$ & $\beta_2$ & $\gamma$ & $\epsilon$ \\
    \hline
    \textit{Qwen3-4B-Instruct} & 0.075 & 0.952 & -0.778 & 0.022 \\
    \textit{Qwen3-8B-Base} & 0.027 & 0.993 & -0.660 & 0.025 \\
    \bottomrule
  \end{tabular} }
\end{table}

\begin{figure}[t]
  \centering
  \includegraphics[width=\columnwidth]{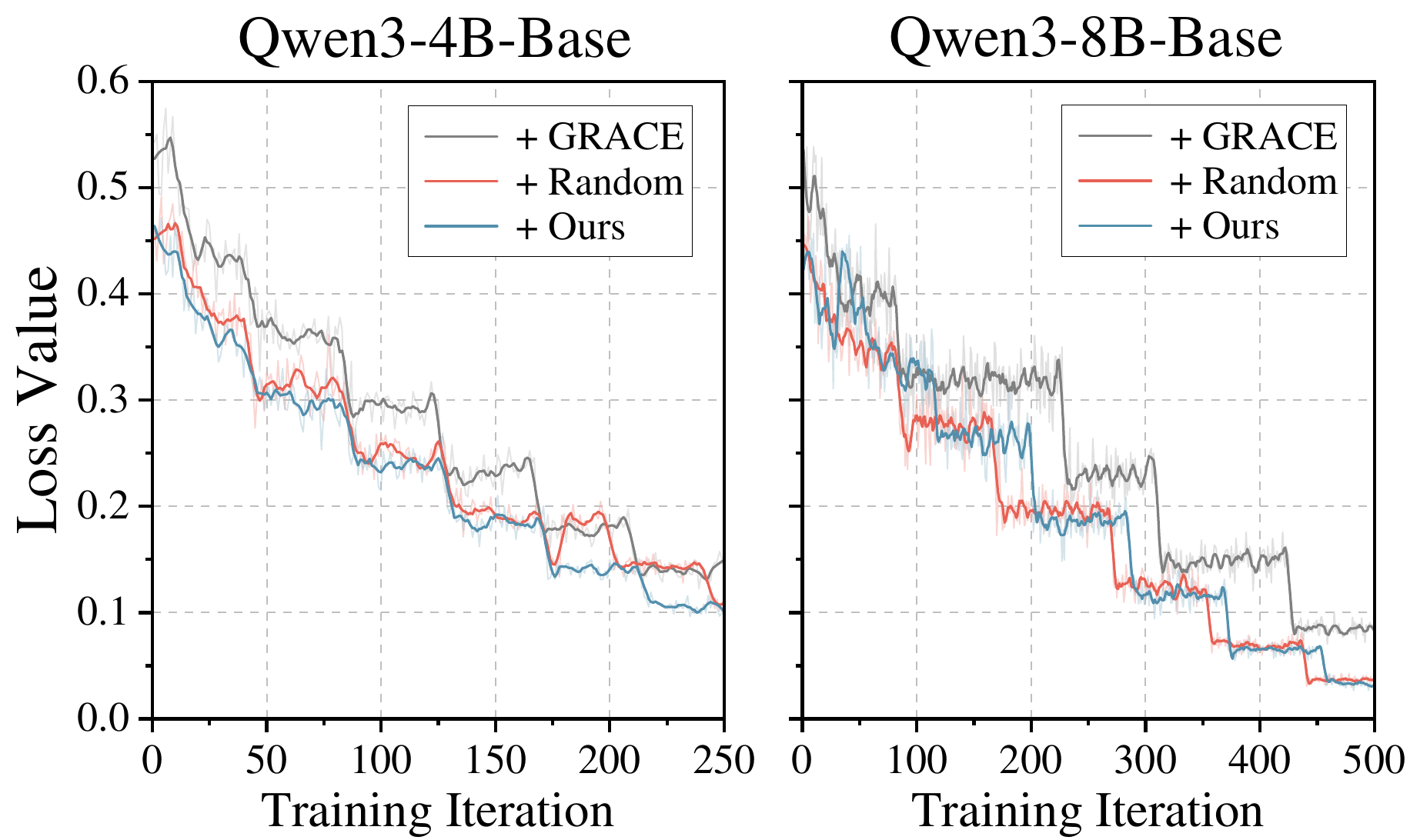}
  \caption{Convergence analysis.}
  \label{convergence}
\end{figure}

\section{More Experimental Results}

\subsection{Convergence Analysis}

In Fig.~\ref{convergence}, we present a convergence analysis showing that the GRACE method, \ie the existing naturalness‑based approach, consistently converges to a higher loss compared with our method. This also demonstrates that our debiasing approach is able to select data with greater naturalness.

\subsection{Comparing More Baselines}

We implement a simple heuristic-based selection method on the same source data as a representative baseline. The results of this comparison are presented in Table~\ref{selection_methods}. On the LIMO-v2 dataset, we compare several baseline selection strategies:  
\begin{itemize}
    \item \textbf{Uniform}: randomly samples 4k reasoning trajectories from the four source LLMs with a uniform distribution to ensure data diversity; 
    \item \textbf{High/Low Difficulty}: selects 4k of the longest or shortest reasoning trajectories, respectively, using response length as a proxy for problem difficulty.
\end{itemize}
The results consistently demonstrate the superior effectiveness of our selected data. Moreover, while more difficult (\ie longer) examples do yield better performance for SFT of reasoning models, confirming that challenging instances are generally more informative, they still contain a higher proportion of redundant or noisy reasoning steps compared to our method. This underscores the advantage of our approach in not only capturing useful difficulty but also filtering out superfluous or low-quality reasoning content.

\begin{table}[t]
\centering
\renewcommand\arraystretch{1.05}
  \caption{Results of more data selection methods.}
  \label{selection_methods}
  \small
  \setlength{\tabcolsep}{5pt}{
  \begin{tabular}{m{2.5cm}<{\centering}m{1.3cm}<{\centering}m{1.3cm}<{\centering}}
    \toprule
    Qwen3-4B-Base & AIME25 & MATH500 \\
    \hline
    + Uniform & 24.16 & 73.20 \\
    + High Difficulty & 26.66 & 77.80 \\
    + Low Difficulty & 20.83 & 71.20 \\
    \rowcolor{lightgrayv} + \babyb & \textbf{30.83} & \textbf{80.00} \\
    \bottomrule
  \end{tabular} }
\end{table}

\subsection{Comparing More Source LLMs}

To further assess generalizability, we also train a non-Qwen model: Llama3-3B, on the data selected by our method \babyb; the results are presented in Table~\ref{more_source}. The results consistently demonstrate the effectiveness of our proposed method.

\begin{table}[t]
\centering
\renewcommand\arraystretch{1.05}
  \caption{Results on Llama3-3B.}
  \label{more_source}
  \small
  \setlength{\tabcolsep}{5pt}{
  \begin{tabular}{m{2.5cm}<{\centering}m{1.3cm}<{\centering}m{1.3cm}<{\centering}}
    \toprule
    Llama3-3B & AIME25 & MATH500 \\
    \hline
    + GRACE & 26.66 & 76.20 \\
    \rowcolor{lightgrayv} + \babyb & \textbf{33.33} & \textbf{84.40} \\
    \bottomrule
  \end{tabular} }
\end{table}